%% file: _main.tex
\input{_constants}
\arxiv 

\pdfoutput=1
\documentclass[10pt,twocolumn,letterpaper]{article}
\input{cvpr_header}
\begin{document}
\title{\paperTitle}
\author{
    Haozhe Jia$^{*1}$ \qquad
    Yan Li$^{*2}$ \qquad
    Hengfei Cui$^2$ \qquad
    Di Xu$^1$ \qquad
    Yuwang Wang$^{\dagger3}$ \qquad
    Tao Yu$^{\dagger3}$ \\
    $^1$Huawei Cloud Computing Technologies Co., Ltd \qquad
    $^2$Northwestern Polytechnical University \\
    $^3$Tsinghua University
    }
%
\twocolumn[{%
\renewcommand\twocolumn[1][]{#1}%
\maketitle

\begin{center}
    \centering
    \captionsetup{type=figure}
    \includegraphics[width=1\textwidth]{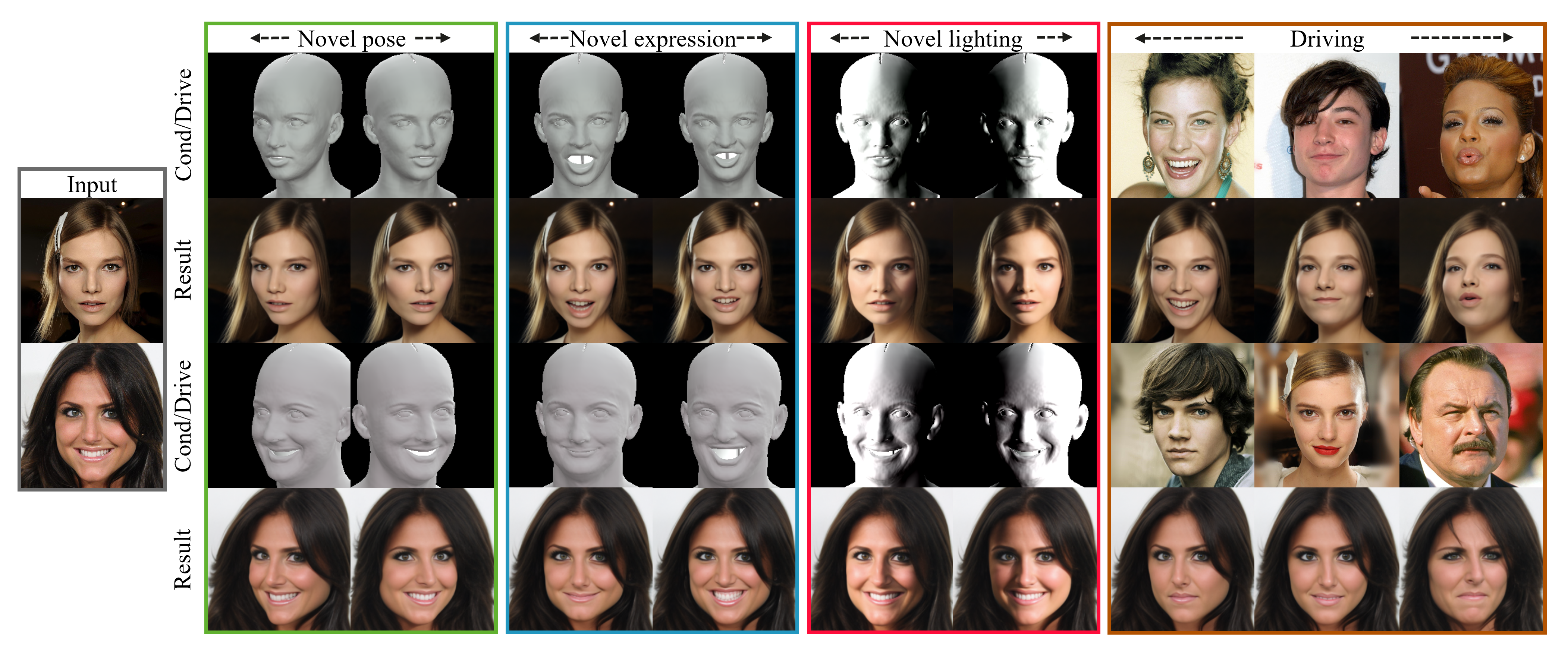}
    \vspace{-0.7cm}
    \captionof{figure}{Our DisControlFace can realistically edit the input portrait corresponding to different explicit parametric controls and faithfully generate personalized facial details of an individual. Our model also supports other editing tasks, \textit{e.g.,} inpainting and semantic manipulations.}
    \label{fig:teaser}
\end{center}%
\vspace{-0.1cm}
}]

\def\thefootnote{$*$}\footnotetext{Equal contributions}
\def\thefootnote{$\dagger$}\footnotetext{Corresponding Authors}

\input{00_abstract}
\input{01_intro}
\input{02_related}
\input{03_prelim}
\input{04_method}
\input{05_exp}
\input{06_conclusion}
\input{07_ackn}

\newpage
{\small
\bibliographystyle{ieee_fullname}
\bibliography{07_references}
}

\ifarxiv \clearpage \input{08_appendix} \fi

\end{document}


\title{\paperTitle \\ Supplemental Material}
\author{\authorBlock}
\maketitle

\appendix
\label{sec:appendix}

\renewcommand\thefigure{\Alph{section}\arabic{figure}}    

\section{Overview}
In this supplement, we present:
\begin{itemize}
    \item Section \ref{implem}: Implementation details, including further descriptions of the training and inference processes as well as detailed architecture of the proposed model.
    \item Section \ref{model_analysis}: Model analysis, including the discussion about the difference between Face-ControlNet and ControlNet, and the validation of the necessity of Masked Diff-AE training.
    \item Section \ref{add_exp}: Additional experiments, including additional ablation study on different inference masking strategies and more visual results.
    \item Section \ref{discuss}: More discussions about limitations and future work.
\end{itemize}

\section{Implementation Details}
\label{implem}
\setcounter{figure}{0}    
\renewcommand\thetable{\Alph{section}\arabic{table}}    
\setcounter{table}{0}
All detailed configurations of generic training and fine-tuning are shown in Table \ref{tab:training_configuration}.
The structural details of DisControlFace is presented in Figure \ref{fig:structure}. 
Specifically, for Diff-AE backbone \cite{preechakul2022diffusion}, we adopted its officially released hyperparameters and pre-trained weights\footnote{\url{https://github.com/phizaz/diffae}.} corresponding to FFHQ256.
Our Face-ControlNet mirrors the structure of the Conditional DDIM (U-Net) in Diff-AE, which however, customizes the input layer by setting the input channel number to 6, allowing it to take the concatenated snapshots as input.
Meanwhile, we fuse the 2D feature maps outputted by the input layers of Conditional DDIM and Face-ControlNet by pixel-wise summation and feed the fused feature maps into the subsequent layers of Face-ControlNet, which tends to improve the quality of the generation images.
Finally, to further provide fine-grained conditioning for the diffusion generation process, we add multi-scale features outputted by the middle and all decoder blocks of the Face-ControlNet ($f_A$ to $f_M$ in Figure \ref{fig:structure}) back to the corresponding blocks of Diff-AE.

\begin{table}
    \centering
    \begin{tabular}{lcc}
    \hline
         &  generic training& fine-tuning\\
         \hline
 Image size& \multicolumn{2}{c}{256}\\
 Patch size&  \multicolumn{2}{c}{16}\\
 Image normalization& \multicolumn{2}{c}{[-1, 1]}\\
 Masking ratio& \multicolumn{2}{c}{sample from $U(0.25, 0.75)$}\\
 Optimizer& \multicolumn{2}{c}{AdamW \cite{loshchilov2018decoupled} (no weight decay)}\\
 Loss     & \multicolumn{2}{c}{MSE loss}\\
 EMA decay& 0.9999&$\setminus$\\
 Learning rate&  1e-4& 1e-5\\
 Batch size&  32& 4\\
 Denoising steps & \multicolumn{2}{c}{1000}\\
 Epochs&  200& 10\\
 Device& 8 V100s&1 V100\\

    \hline
    \end{tabular}
    \caption{\textbf{Training and fine-tuning configurations.}}
    \label{tab:training_configuration}
\end{table}
\input{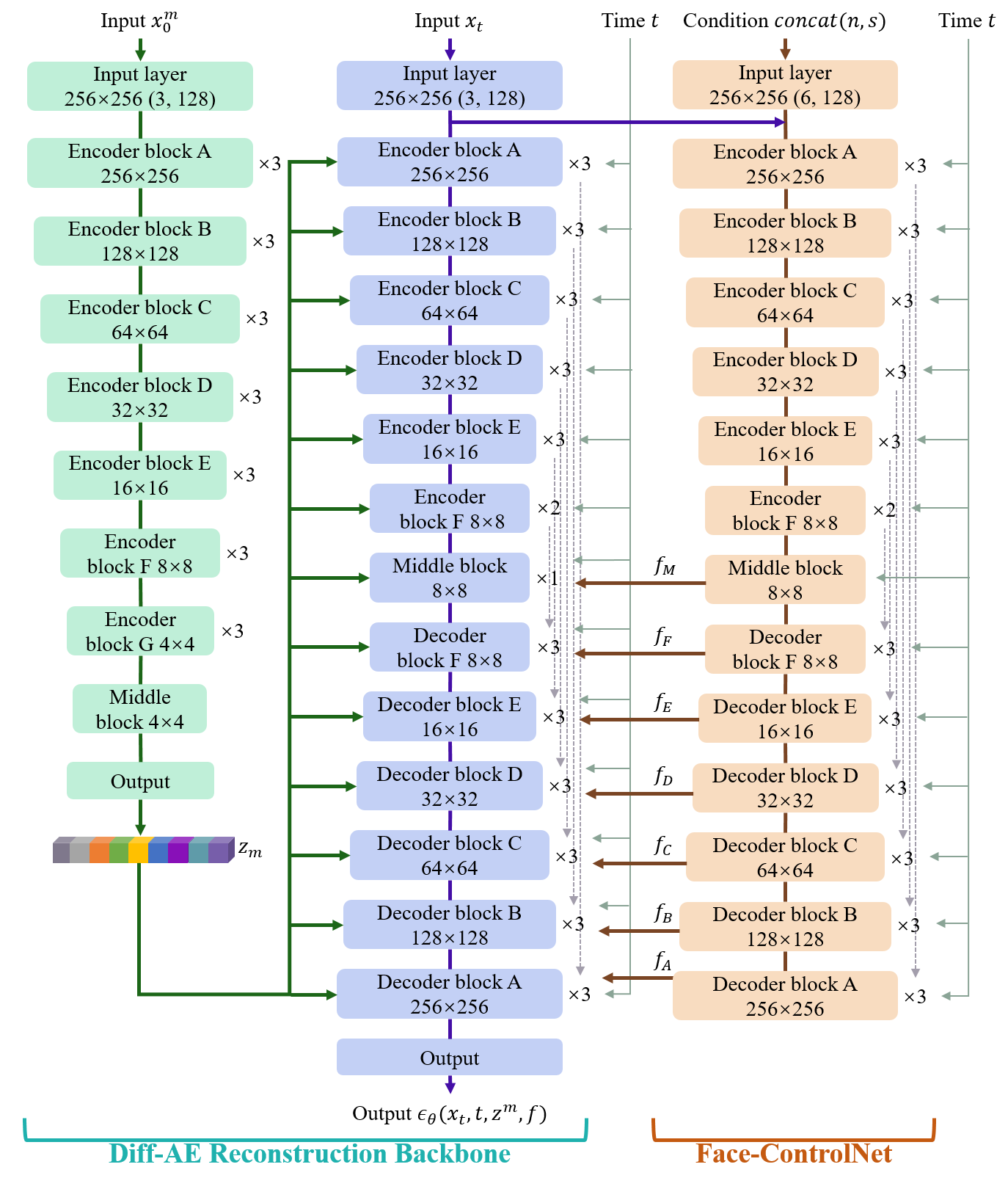}

\section{Model Analysis}
\label{model_analysis}
\subsection{Face-ControlNet v.s ControlNet}
\input{figs/controlnet_example}
Our Face-ControlNet is inspired from ControlNet \cite{zhang2023adding}, a popular deep network which has been widely employed to add various spatial visual guidance (\textit{e.g.,} edge maps, pose maps, depth maps, \textit{etc}.) to Stable Diffusion (SD) \cite{rombach2022high} for text-to-image generation.
However, there exists many differences between two models. 
First, ControlNet is specially designed for SD which can generate specific content based on the input prompts, while there still exists uncertainty and randomness in the generation controlled by visual guidance and text prompts.
As shown in Figure \ref{fig:controlnet_example}, given a specific inference prompt and a HED Boundary map of a portrait image, SD combined with ControlNet generates images matching the input HED Boundary but with diverse details and semantics.
In contrast to this, this paper focuses on reconstruction and editing of the input portrait image.
Given this, our Face-ControlNet is designed to be compatible with a deterministic reconstruction backbone, aiming to better address facial image editing while recovering all edit-unrelated details.
Also due to this, different from ControlNet, it is infeasible to train Face-ControlNet with the traditional training strategy of diffusion model.
Second, we construct Face-ControlNet as a U-Net instead of adopting zero convolutions introduced in ControlNet.
With this setting, we can endow Face-ControlNet with strong capability of U-shape models in extracting spatial deep features from pixel-aligned visual guidance map (two snapshots in this work).
In practice, we find that constructing deep decoder indeed can obtain better generation quality than using zero convolutions.

\subsection{Effectiveness of Masked Diff-AE training}
\input{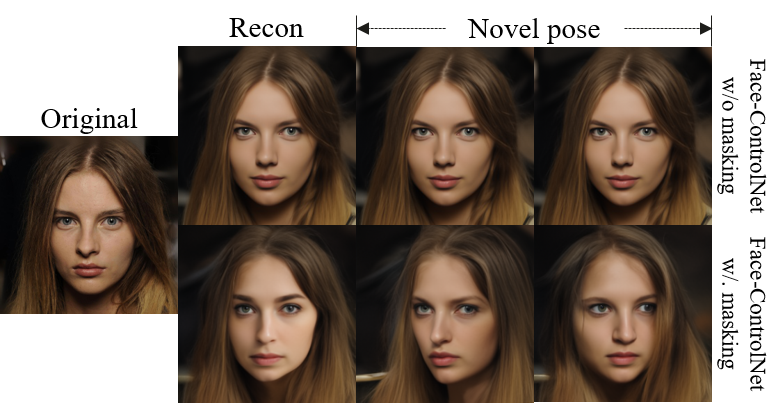}
In this paper, we achieve a disentangled training of the proposed Face-ControlNet by reformulating the traditional training pipeline as a masked diffusion autoencoder (Masked Diff-AE).
To demonstrate the necessity of the proposed Maksed Diff-AE training, we separately train Face-ControlNet with and without random masking for the input image of semantic encoder of the Diff-AE backbone.  
Figure \ref{Masked-Diff-AE} shows that both training strategies enable the model to reconstruct the input image. 
However, only the model trained in the form of Masked Diff-AE can generate images with novel poses. 
This result supports the point we mentioned in the main paper that only limited gradients can be generated during error back-propagation since the pre-trained Diff-AE backbone can already allow near-exact image reconstruction, which are far from sufficient to effectively train Face-ControlNet.
By means of our Masked Diff-AE training, the semantic encoder of Diff-AE can generate the semantic latent code containing fragmented and incomplete content and spatial information of the input image, which facilitates Face-ControlNet to recover the masked face regions and thereby learning explicit face control.

\section{Additional Experiments}
\label{add_exp}
\subsection{Additional Ablation Studies}

\vspace{0.8em}\noindent\textbf{Different masking strategies in inference.}\quad
Here we explore how the masking strategy used in inference affects  the generation performance.
The comparison is shown in Figure \ref{fig:ab5}, we can observe that when setting masking ratio to 25\% for all inference steps, the edited image can not match the control signal very well which can be attributed to strong deterministic reconstruction in this setting.
Meanwhile, setting masking ratio to 75\% for all inference steps slightly harm the semantics recovering.
Besides, we can see the other three masking strategies can achieve better editing results where the proposed linear masking ratio can perform overall best editing with accurate face control and good preservation of facial semantics.

\input{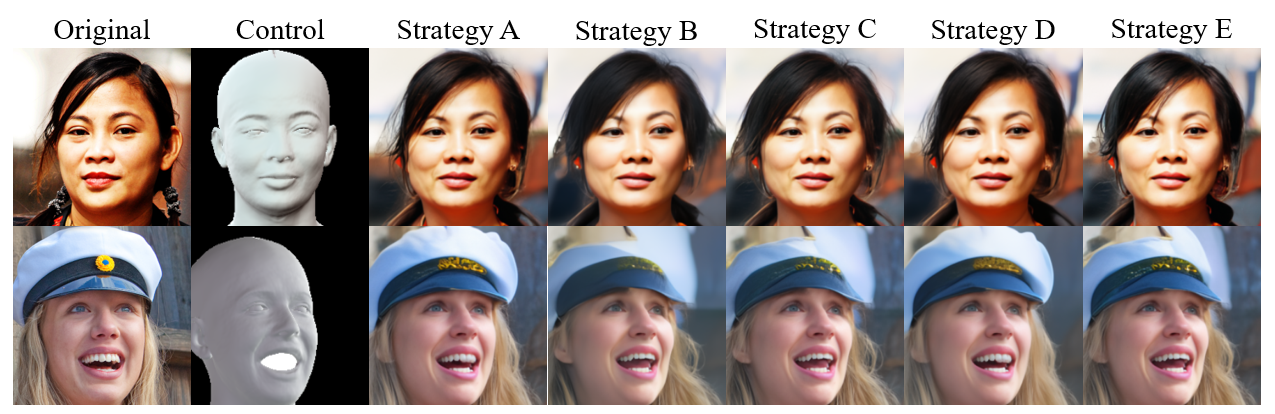}
\input{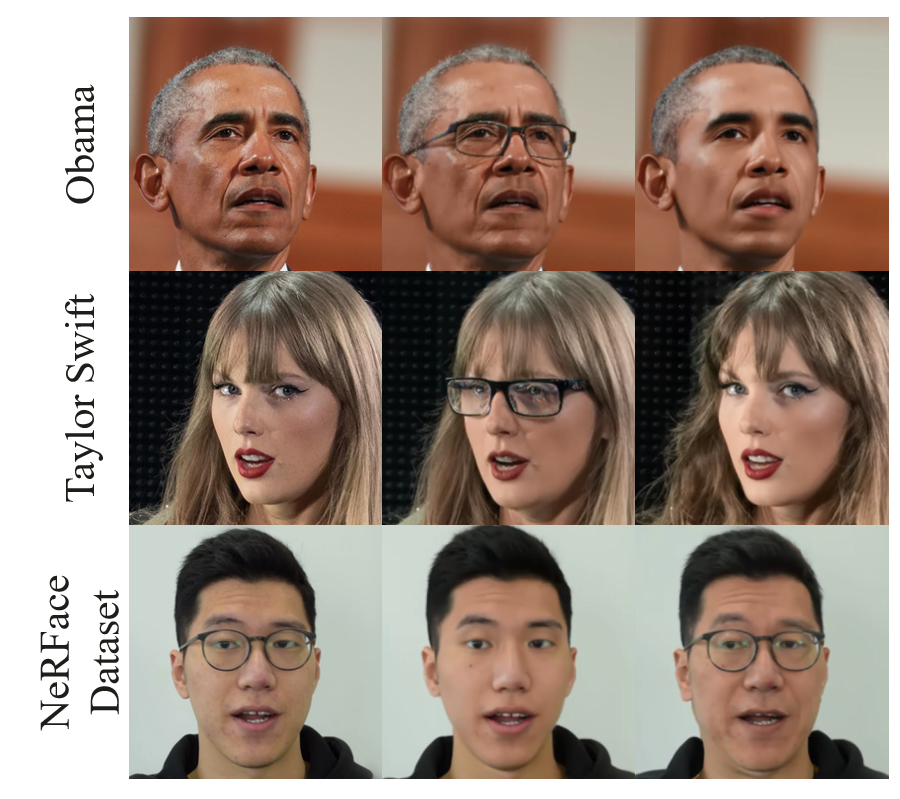}
\input{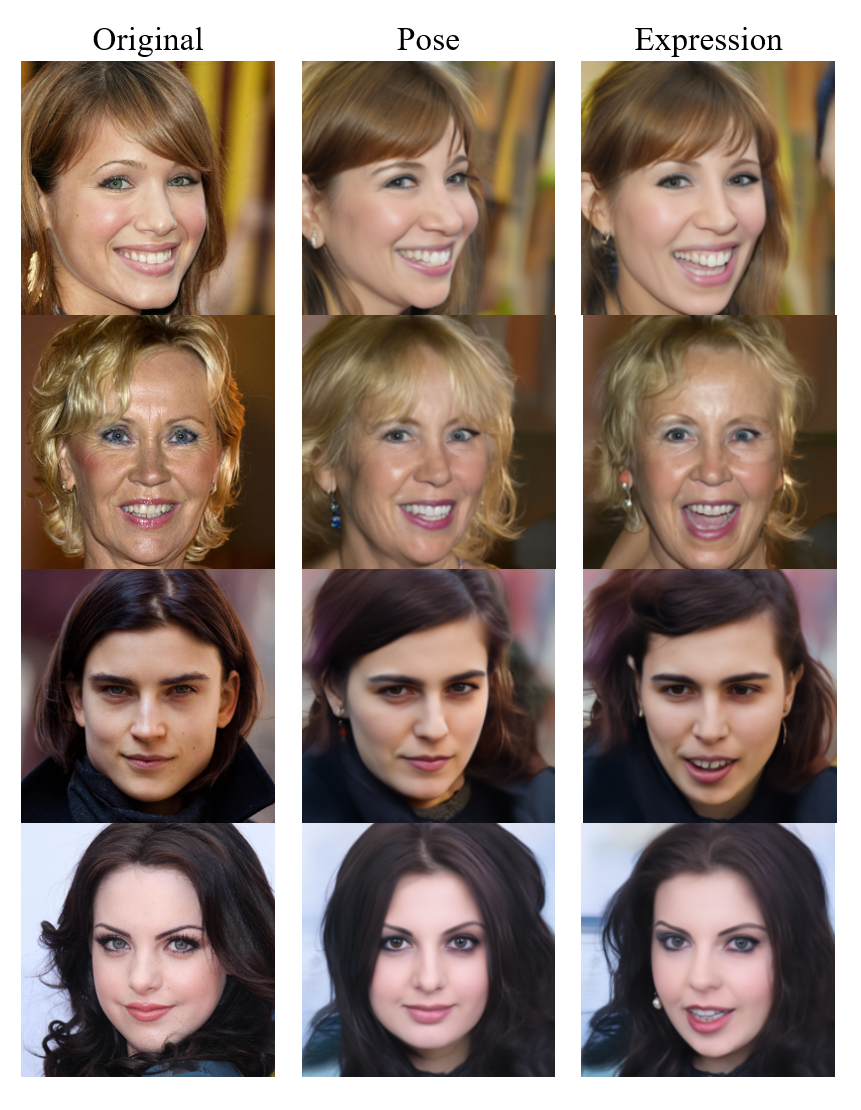}
\input{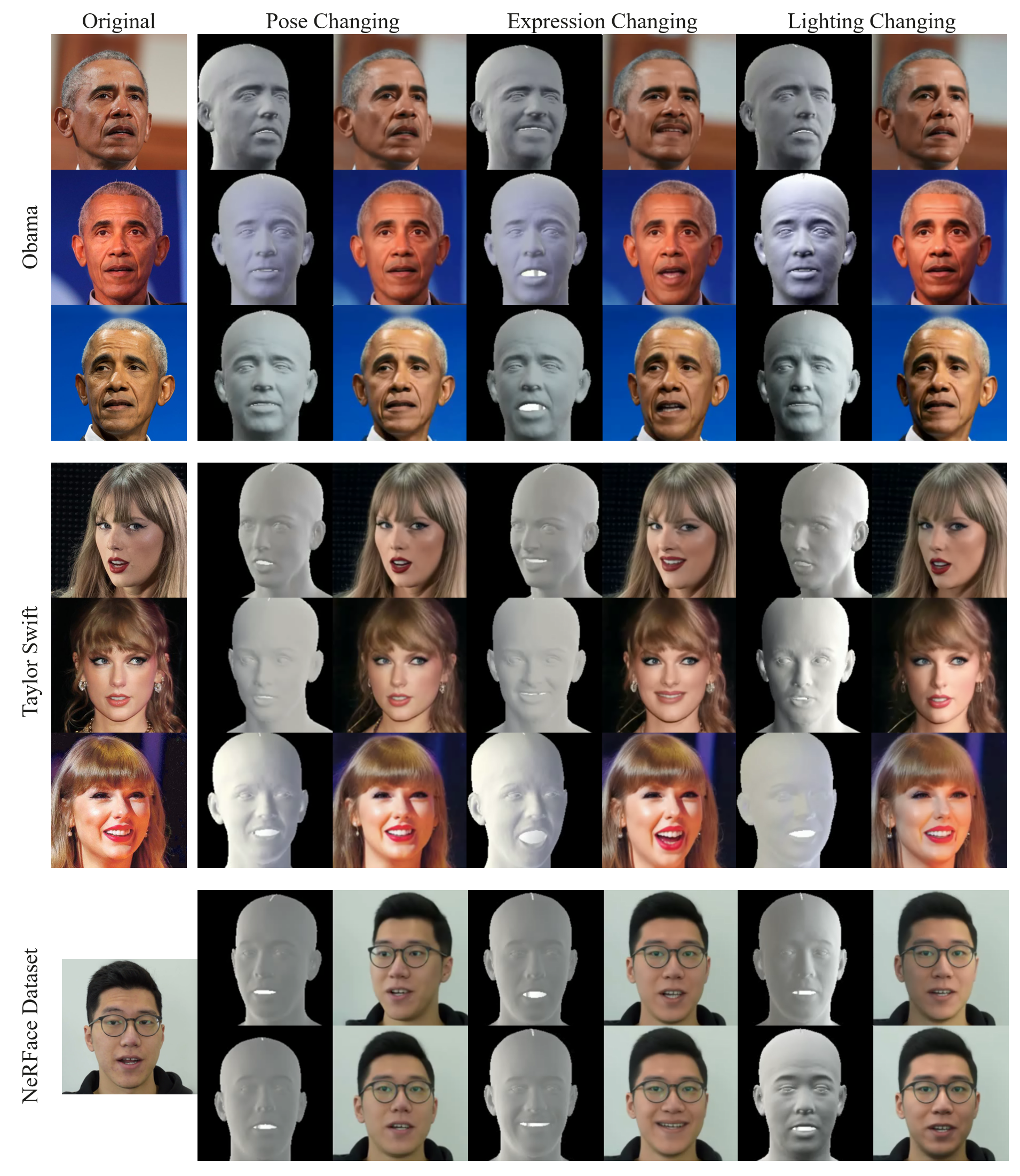}
\subsection{More Visual Results}
In this section, we provide additional visual results of our DisControlFace:
\begin{itemize}
    \item Figure \ref{fig:few-shot-manipulation} visualizes semantic manipulations on Obama, Taylor, and a person in NeRFace dataset. 
    \item Figure \ref{fig:zero-shot-ofd} provides zero-shot editing results on CeleA-HQ dataset \cite{karras2018progressive}.
    \item Figure \ref{fig:more-few-shot} shows more few-shot editing results.
\end{itemize}

\section{Additional Limitations and Future Work}
\label{discuss}
it would be more reasonable to perform random masking only on the face-related patches
Our DisControlFace has a parallel editing control network besides the U-Net noise predictor and performs denoising diffusion process in image space, which results in the model being able to generate images with limited resolutions.
Potential future improvements includes introducing a light-weight super-resolution network to the model or extending the model to a latent diffusion version.

In the proposed Masked Diff-AE training, we randomly mask some patches of the whole image and take the masked image as the input of the semantic encoder of Diff-AE backbone to achieve an effective disentangled training of Face-ControlNet.
Meanwhile, only performing random masking on face-related patches is expected to further improve the consistency of the background region of the input portrait with different editing applied, which however might slightly increase the training and inference time.
Corresponding explorations and experiments can serve as another potential valuable future work.

\clearpage
{\small
\bibliographystyle{ieee_fullname}
\bibliography{09_supp_refs}
}

%% file: _constants.tex
\def\cvprPaperID{4919}
\def\confName{CVPR}
\def\confYear{2024}

\def\paperTitle{DisControlFace: Adding Disentangled Control to Diffusion Autoencoder for One-shot Explicit Facial Image Editing}

\newif\ifreview 
\newif\ifarxiv \newcommand{\arxiv}{\arxivtrue}
\newif\ifcamera 
\newif\ifrebuttal 

%% file: cvpr_header.tex
\ifreview \usepackage[review]{cvpr} \fi
\ifarxiv \usepackage[pagenumbers]{cvpr} \fi
\ifrebuttal \usepackage[rebuttal]{cvpr} \fi
\ifcamera \usepackage{cvpr} \fi

\usepackage{graphicx}
\usepackage{amsmath}
\usepackage{amssymb}
\usepackage{booktabs}

\input{_macros}  

\usepackage{xr-hyper}

\makeatletter
\newcommand*{\addFileDependency}[1]{
  \typeout{(#1)}
  \@addtofilelist{#1}
  \IfFileExists{#1}{}{\typeout{No file #1.}}
}

\makeatother

\usepackage[pagebackref,breaklinks,colorlinks]{hyperref}
\usepackage[capitalize]{cleveref}
\crefname{section}{Sec.}{Secs.}
\crefname{table}{Table}{Tables}
\crefname{figure}{Fig.}{Figs.}

%% file: _macros.tex
\usepackage{times}
\usepackage{microtype}
\usepackage{epsfig}
\usepackage[table,xcdraw]{xcolor}
\usepackage{caption}
\usepackage{float}
\usepackage{placeins}
\usepackage{color, colortbl}
\usepackage{stfloats}
\usepackage{enumitem}
\usepackage{tabularx}
\usepackage{xstring}
\usepackage{multirow}
\usepackage{xspace}
\usepackage{url}
\usepackage{subcaption}
\usepackage{xcolor}
\usepackage[hang,flushmargin]{footmisc}

\ifcamera \usepackage[accsupp]{axessibility} \fi

% ...

\ifarxiv  \fi

\newcommand{\R}[1]{{%
    \textbf{%
        \ifstrequal{#1}{1}{\textcolor{red}{R#1}}{%
        \ifstrequal{#1}{2}{\textcolor{blue}{R#1}}{%
        \ifstrequal{#1}{3}{\textcolor{magenta}{R#1}}{%
        \ifstrequal{#1}{4}{\textcolor{teal}{R#1}}{%
                           \textcolor{cyan}{R#1}%
        }}}}%
    }%
}}

%% file: 00_abstract.tex
\begin{abstract}
In this work, we focus on exploring explicit fine-grained control of generative facial image editing, all while generating faithful facial appearances and consistent semantic details, which however, is quite challenging and has not been extensively explored, especially under an one-shot scenario.
We identify the key challenge as the exploration of disentangled conditional control between high-level semantics and explicit parameters (\textit{e.g.,} 3DMM) in the generation process, and accordingly propose a novel diffusion-based editing framework, named DisControlFace.
Specifically, we leverage a Diffusion Autoencoder (Diff-AE) as the semantic reconstruction backbone. 
To enable explicit face editing, we construct an Exp-FaceNet that is compatible with Diff-AE to generate spatial-wise explicit control conditions based on estimated 3DMM parameters.
Different from current diffusion-based editing methods that train the whole conditional generative model from scratch, we freeze the pre-trained weights of the Diff-AE to maintain its semantically deterministic conditioning capability and accordingly propose a random semantic masking (RSM) strategy to effectively achieve an independent training of Exp-FaceNet.
This setting endows the model with disentangled face control meanwhile reducing semantic information shift in editing.
Our model can be trained using 2D in-the-wild portrait images without requiring 3D or video data and perform robust editing on any new facial image through a simple one-shot fine-tuning.
Comprehensive experiments demonstrate that DisControlFace can generate realistic facial images with better editing accuracy and identity preservation over state-of-the-art methods. 
Project page: \href{https://discontrolface.github.io/}{https://discontrolface.github.io/}
\end{abstract}

%% file: 01_intro.tex
\section{Introduction}
\label{sec:intro}
Facial image editing has long been a hot research topic in the fields of computer vision and computer graphics, where the key challenge is to effectively achieve fine-grained controllable generation of realistic facial images while preserving semantic face priors.

3D Morphable Models (3DMMs) \cite{blanz2023morphable,gerig2018morphable} have been widely employed to represent variations in facial shape and texture \cite{cao2013facewarehouse,li2017learning,gerig2018morphable,gupta2010texas,tewari2019fml,tewari2021learning,tran2019towards,tran2019learning}, whereas their ability to capture personalized facial features is limited and the performance highly depends on the quality and diversity of  the 3D face training data.
On top of this, subsequent learning-based explicit face models \cite{sela2017unrestricted,yang2020facescape,feng2021learning,danvevcek2022emoca,lei2023hierarchical,dib2021towards,tewari2019fml,tewari2021learning,tran2019towards,tran2019learning} achieve controllable generation of dynamic and expressive facial animations by capturing the nuances of facial features under different expressions, poses, and lighting conditions, nevertheless, can neither generate realistic facial appearances that correspond to the animated 3D face geometries nor model refined geometric details in non-facial regions, \textit{e.g.,} hair, eyes, and mouth.
Follow-up efforts integrate explicit facial modeling with implicit 3D-aware representations like Neural Radiance Fields (NeRFs) to reconstruct animated realistic head avatars \cite{gafni2021dynamic,zheng2022avatar,peng2021animatable,liu2021neural,park2021nerfies,hong2022headnerf,peng2020convolutional,mildenhall2020nerf,mescheder2019occupancy,park2019deepsdf,sitzmann2020implicit,sitzmann2019deepvoxels}, which however, heavily rely on 3D consistent data such as monocular portrait videos and tend to exhibit limited generalization.

In contrast, generative face models enable single image reconstruction and editing due to the superior capability in learning rich face priors from in-the-wild portraits.
Recent GAN-based approaches \cite{chan2021pi,chan2022efficient,deng2022gram,gu2021stylenerf,bergman2022generative,hong2022eva3d,noguchi2022unsupervised,sun2022controllable,wu2022anifacegan} incorporate explicit 3D facial priors and implicit neural representations to achieve directed generation of high-resolution, realistic, and view-consistent facial images without the need of 3D face scans or portrait videos.
However, those methods mainly provide implicit or limited explicit controls of face generation.

More recently, the diffusion-based framework \cite{ho2020denoising} has emerged as the predominant choice for various generation tasks, owing to its impressive performance and diverse conditioning options.
Some approaches have also shown promising enhancements in face reconstruction and various face editing tasks, such as face relighting \cite{ponglertnapakorn2023difareli}, semantic attributes manipulation \cite{preechakul2022diffusion}, and explicit appearance control \cite{ding2023diffusionrig}.
Unfortunately, it can be seen that when editing and modifying some specific facial attributes, other facial attributes or editing-irrelevant details often occur unexpected and uncontrollable changes, leading to an incoherent and identity-altered generated face. 
This prevalent issue can be attributed to that these generative face models struggle to effectively perform disentangled control in the generation process.

In this work, we propose a novel diffusion-based generative framework, namely DisControlFace to achieve one-shot editing of facial images.
To generate a photo-realistic, high-fidelity facial appearance corresponding to specific explicit parameters (\textit{e.g.,} 3DMM) while faithfully preserving high-level semantic priors, we particularly focus on enhancing the diffusion model with disentangled conditional control.
Specifically, we adopt a Diffusion Autoencoder (Diff-AE) \cite{preechakul2022diffusion} as the generative backbone, which can enable a deterministic image reconstruction by conditioning Denoising Diffusion Implicit Model (DDIM)\cite{song2020denoising} on the semantic information of the input image.
We then specially construct an Exp-FaceNet compatible with the Diff-AE backbone, which further provides multi-scale, spatial-aware DDIM conditioning corresponding to the facial parameters of shape, pose, expression, and lighting.
Moreover, we claim that training different DDIM conditioning together, as with existing methods is not conductive to learning disentangled face control.
We therefore freeze the pre-trained weights of Diff-AE and accordingly design a random semantic masking (RSM) strategy to enable the training of Exp-FaceNet, by means of which the model can learn explicit parametric face control independently without affecting semantically deterministic DDIM conditioning.
Also benefiting from this disentangled setting, instead of relying on 3D or video data, we can utilize 2D in-the-wild portrait dataset such as FFHQ \cite{karras2019style} to effectively train Exp-FaceNet to learn a robust and generalized capability in explicit face editing.
Considering there exists domain gap between the pre-trained face data and the target new face image, which tends to prevent Diff-AE from performing near-exact semantic reconstruction, we finally introduce an one-shot fine-tuning to Diff-AE so as to restore personal identity and editing-irrelevant details of the input portrait under a subject-agnostic scenario. 
Our approach not only achieves state-of-the-art (SOTA) qualitative and quantitative results for one-shot explicit facial image editing, but also supports generating realistic and faithful facial appearance of specific individuals in image inpainting, semantic attributes manipulations, and cross-identity face driving (shown in Figure \ref{fig:teaser}).

Our contributions can be summarized as follows:
\begin{itemize}
    \item We propose a novel diffusion-based generation framework, consisting of a Diffusion Autoencoder (Diff-AE) backbone and an explicit face control network (Exp-FaceNet) for synthesizing photo-realistic, high-fidelity portrait images corresponding to the editing of explicit facial properties only trained with 2D in-the-wild images. 
    \item To the best of our knowledge, we are the first to introduce a weight-frozen pre-trained Diff-AE to explicit face editing pipeline to provide deterministic semantic conditioning, meanwhile designing an effective training strategy to enable the Exp-FaceNet with a disentangled explicit parametric (\textit{e.g.,} 3DMM) conditioning.
    \item Our method achieves SOTA generation performance in explicit facial image editing, and also supports various one-shot face editing tasks.
\end{itemize}

%% file: 02_related.tex
\section{Related Work}
\label{sec:related}
\noindent\textbf{Generative Face Modeling.} \quad  Various GAN-based models \cite{nguyen2019hologan,karras2019style,karras2020analyzing,schwarz2020graf,chan2021pi,pan2021shading} have been proposed to synthesize realistic facial images by learning the underlying data priors from large-scale in-the-wild images \cite{schwarz2020graf,chan2021pi,pan2021shading}.
However, when it comes to precise control and interpretability, those generative face models fall short compared to explicit parametric models.
Given this, several approaches \cite{chan2022efficient,sun2022controllable,wu2022anifacegan,sun2023next3d} go a step further by integrating explicit parameters and GANs, which can simultaneously generate highly realistic and coherent portraits and achieve fine-grained control of facial appearance.
Recently, diffusion models \cite{ho2020denoising,song2020denoising}  have gained recognition for their superior ability to learn data distributions compared to GANs, and thus have been widely adopted to generate realistic and diverse images in various generation tasks, including facial image synthesis \cite{preechakul2022diffusion,preechakul2022diffusion}.
DiffusionRig \cite{ding2023diffusionrig}, a closely related method, introduces pixel-aligned physical properties rendered from explicit parameters estimated by DECA \cite{feng2021learning} to denoising diffusion process to generate photo-realistic facial images corresponding to target pose, expression, and lighting conditions, which however, highly relies on a personalized fine-tuning (around 20 images) to preserve the facial appearance priors of a specific person .
Similar problems widely exist in most conditional diffusion face models, as can be observed where identity shifts and unexpected attributes alterations may occur during face reconstruction and editing.
This can be attributed to the lack of disentangled control capabilities when conditioning diffusion models with both facial semantics and physical information.
DisControlFace overcomes this challenge by leveraging a weight-frozen pre-trained Diff-AE to provide semantically deterministic DDIM conditioning and independently training a separate Exp-FaceNet to learn disentangled face control based on 3DMM parameters.

\noindent\textbf{Conditional Diffusion Model.}\quad 
Conditional DDIM enables the Denoising Diffusion Probabilistic Model (DDPM) \cite{ho2020denoising} to generate content consistent with specific control signals through various conditioning manners.
Most existing models encode various control information into global conditional vectors, which can be text embedding \cite{rombach2022high} or semantic embedding \cite{preechakul2022diffusion,ponglertnapakorn2023difareli}.
To achieve spatial-aware and precise control of the generation, some approaches (\textit{e.g.,} DiffusionRig \cite{ding2023diffusionrig} and SR3 \cite{saharia2022image} ) concatenate various spatial conditions and denoised images together as the input of the U-Net noise predictor in each denoising step.
However, this form of conditioning requires the U-Net to have a unique input layer, resulting in the model having limited generalization and making it hard to reuse existing well-trained diffusion models.
Besides, some other methods specially construct spatial conditioning branches to extract spatial-aligned conditional features and insert them into the U-Net \cite{zhang2023adding,wang2023disco,ponglertnapakorn2023difareli}.
ControlNet \cite{zhang2023adding} has been widely employed to add various spatial visual guidance (\textit{e.g.,} edge maps, pose maps, depth maps, \textit{etc}.) to text-to-image generation models such as Stable Diffusion \cite{rombach2022high}.
Whereas, it may not suitable for deterministic reconstruction or editing tasks since there still exists uncertainty and randomness in the generation controlled by visual guidance and text prompts.
In contrast, our Exp-FaceNet is specially designed to be compatible with a semantic reconstruction DDIM, Diff-AE \cite{preechakul2022diffusion}, aiming to better address facial image editing based on explicit control information.

\noindent\textbf{Learning Specific Facial Priors.}\quad Effectively extracting the facial appearance priors of the specific person and injecting this global prior into the generation process is crucial for preserving facial semantics such as identity, accessories, hairstyle, and background information in facial image editing.
Most existing generative methods address this issue by fine-tuning the network in various settings \cite{nitzan2022mystyle,hu2021lora,gal2023encoder,ruiz2023dreambooth}, or designing special optimization strategies, such as identity penalty, face recognition loss, and latent representation editing \cite{liu20223d,bansal2023universal,yuan2023inserting,valevski2023face0}.
In contrast to previous work like \cite{ding2023diffusionrig,nitzan2022mystyle} which collect personal albums to learn personalized facial priors, in this work, we focus on the subject-agnostic editing scenario, which is more challenging but practical.
On the basis of the inherent and maintained facial semantics capturing capability of Diff-AE, only a fast and yet simple fine-tuning using an out-of-domain new face image is needed to faithfully restore the semantic appearance details of the target image during editing.

%% file: 03_prelim.tex
\section{Preliminaries}
\label{sec:preliminaries} 
\subsection{3D Morphable Face Models}
FLAME \cite{li2017learning} , a popular 3DMM model, can be expressed as $ M(\boldsymbol{\beta} , \boldsymbol{\theta} , \boldsymbol{\psi}  ):\mathbb{R}^{\lvert \boldsymbol{\beta}  \rvert \times \lvert \boldsymbol{\theta}  \rvert \times \lvert \boldsymbol{\psi} \rvert }  \rightarrow \mathbb{R}^{3N} $, which takes shape $\boldsymbol{\beta}$, pose $\boldsymbol{\theta} $, and expression $\boldsymbol{\psi} $ as inputs and outputs a face mesh with $N$ vertices.
On this basis, some off-the-shelf 3D face estimators, such as DECA \cite{feng2021learning} and EMOCA \cite{danvevcek2022emoca}, achieve 3D face reconstruction by regressing individual-specific FLAME parameters from in-the-wild images.
We utilize EMOCA to obtain a face mesh corresponding to the input image with enhanced expression consistency, and render it to pixel-aligned explicit conditions.
Specifically, given a single image $I$, the coarse branch of EMOCA estimate its corresponding $\boldsymbol{\beta}$, $\boldsymbol{\theta}$, $\boldsymbol{\psi}$, albedo $\boldsymbol{\alpha}$, spherical harmonic (SH) illumination coefficient $\boldsymbol{l}$, and camera $c$, which can be expressed as $E_{c} (I) \rightarrow (\boldsymbol{\beta} , \boldsymbol{\theta} , \boldsymbol{\psi}, \boldsymbol{\alpha}, \boldsymbol{l}, \boldsymbol{c})$. 
The detailed branch further outputs a detail vector $\delta $ and computes the displacement map $D$ in UV space, which is specifically expressed as $E_{d}(I) \rightarrow \boldsymbol{\delta}$ and $F_{d} (\boldsymbol{\delta} , \boldsymbol{\psi} , \boldsymbol{\theta} _{jaw} )\rightarrow D$. 

\input{figs/pipeline}
\subsection{Diffusion Autoencoders (Diff-AE)}
\label{subsec:diffae} 
Diff-AE \cite{preechakul2022diffusion} reformulates the traditional diffusion generation model into an autoencoder and captures high-level image semantics for DDIM conditioning, therefore supporting near-exact reconstruction and attribute manipulation of the input image.
Specifically, Diff-AE uses a semantic encoder to generate a 512-dimensional latent code $z$ with global semantics of the input image $x_{0}$.
Then, $z$ can be introduced to the reverse deterministic generative process of DDIM to obtain a noisy map $x_{T}^{}$ which captures the stochastic variations of $x_{0}$.
Last, a conditional DDIM model decodes $(z, x_{T}^{})$  to achieve a deterministic reconstruction of $x_{0}$. 
By linearly modifying the semantic latent code $z$, Diff-AE can manipulate diverse global semantic attributes, \textit{e.g.,} age, gender, and hairstyle.
We introduce Diff-AE to our DisControlFace as the reconstruction backbone and freeze its pre-trained weights to provide semantically deterministic DDIM conditioning during the training of explicit face control.

%% file: figs/pipeline.tex
\begin{figure*}[t]
    \centering
    \includegraphics[width=\linewidth]{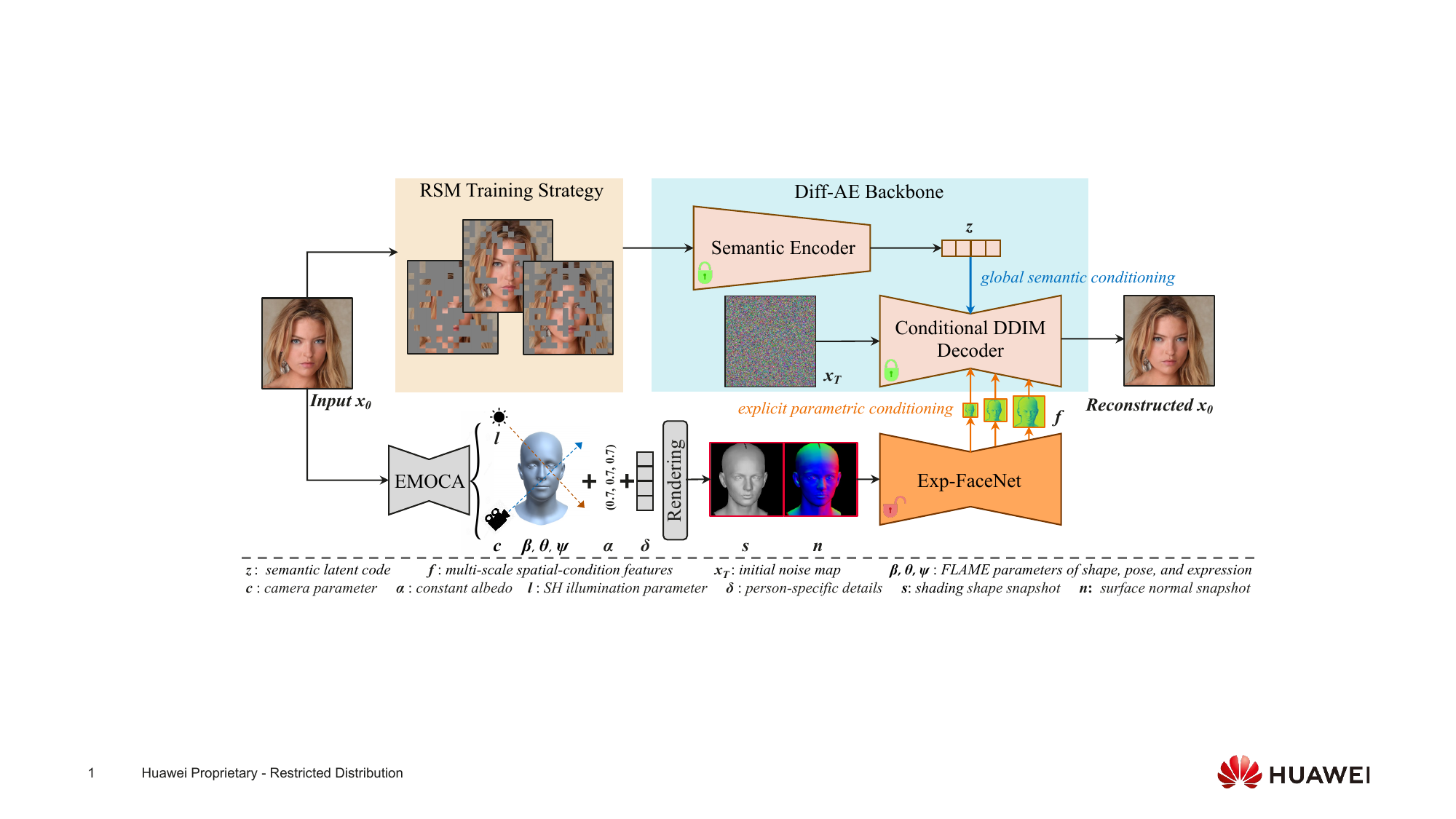}   \caption{\textbf{Pipeline overview.} Our DisControlFace leverages Diffusion Autoencoder (Diff-AE) as the reconstruction backbone freeze its pre-trained weights to maintain the semantic deterministic conditioning capability, which is effective in reducing semantic information shift during the editing of the input portrait image. Then, an explicit face control network, Exp-FaceNet compatible with the Diff-AE is constructed, which takes pixel-aligned snapshots rendered from estimated explicit parameters as inputs and generates multi-scale control features to condition the DDIM decoder.  
    Moreover, a random semantic masking (RSM) training strategy is accordingly designed to enable a disentangled explicit face control of Exp-FaceNet.} 
    \label{fig:pipeline}
\end{figure*}

%% file: 04_method.tex
\section{Method}
\label{sec:Method}
In the pursuit of a robust one-shot explicit facial image editing, we propose a generative framework, namely DisControlFace, providing DDIM conditioning on disentangled face control between high-level semantics and explicit 3DMM parameters (shown in Figure \ref{fig:pipeline}).
Specifically, we adopt a weight-frozen Diff-AE as a semantic reconstruction backbone and construct an Exp-FaceNet to provide explicit parametric face control (Sec. \ref{subsec:Exp-FaceNet}).
Furthermore, we design a training strategy to effectively enable the training of Exp-FaceNet (Sec. \ref{subsec:training strategy}).
Finally, we adopt an one-shot fine-tuning to improve the semantic consistency and faithfulness of the generated facial image under the subject-agnostic editing scenario (Sec. \ref{subsec:one-shot finetuning}).


\subsection{Exp-FaceNet}
\label{subsec:Exp-FaceNet}
On the basis of the adopted Diff-AE reconstruction backbone, an intuitive idea for learning explicit face editing capability is to further build additional DDIM conditioning on 3DMM parameters.
Compared to directly adopting the 3DMM parameters as non-spatial control conditions, generating pixel-aligned conditional maps based on those parameters is more conducive and compatible to convolution-based visual representation, which also helps enable fine-grained spatial control for the denoising diffusion process.
Given this, we specially construct Exp-FaceNet, an explicit control network compatible with the adopted Diff-AE reconstruction backbone to perform a disentangled spatial conditioning for DDIM-based generation.
Here we first estimate 3DMM parameters from facial images and transfer them to the corresponding visual guidance map.
Specifically, we use EMOCA \cite{danvevcek2022emoca} to predict FLAME parameters (including shape $\beta$, pose $\theta$, and expression $\psi$), SH illumination parameter $l$, and camera parameter $c$ from an input portrait.
Different from those previous work \cite{ding2023diffusionrig,ghosh2020gif,ponglertnapakorn2023difareli}, 
here we also adopt the person-specific detail vector $\delta$ estimated by EMOCA, which can be combined with $\theta$ and $\psi$ to generate the expression-dependent displacement map for refining the face geometry with animatable wrinkle details.
To avoid undesired disturbance to facial appearance priors caused by inaccurate and unrealistic appearance estimation, we set the albedo map $\alpha$ to a constant gray value, thereby focusing on controlling the edit of shape, pose, expression, and lighting.
We render these explicit parameters into a surface normal snapshot $n$: 
\begin{equation}
n=\mathcal{R}(\mathcal{G}(\mathcal{M}(\beta, \theta, \psi)), c, \delta)
\end{equation}
where the FLAME model $\mathcal{M}$ is used to calculate the 3D face mesh, $\mathcal{G}$ and $\mathcal{R}$  indicate normal calculation function and the Lambertian reflectance renderer, respectively.
By means of this, $n$ can reflect the fine-grained geometry of the input face and is compatible with pose parameter $\theta$ and expression parameter $\psi$.
Furthermore, we also generate a shading shape snapshot $s$ to illustrate the lighting conditions associated with the SH illumination parameter $l$:
\begin{equation}
s=\mathcal{R}(\mathcal{M}(\beta, \theta, \psi), \alpha, l, c, \delta)
\end{equation}
Considering U-Net \cite{ronneberger2015u} excels at extracting spatial features from images for various vision tasks, we construct Exp-FaceNet in a similar U-shape structure, which takes channel-concatenated snapshots $n$ and $s$ as the input visual guidance map and generates multi-scale deep features for spatial-aware conditioning.
Then we feed back the spatial-condition features outputted by each stage of the U-Net's decoder back to Diff-AE to provide multi-scale conditional control.
Please see the detailed architecture in the supplement.

\input{figs/one_shot}
\subsection{RSM Training Strategy}
\label{subsec:training strategy}
DisControlFace can be regarded as a DDIM model that is simultaneously conditioned by a global semantic code and multi-scale explicit-control feature maps (see Figure \ref{fig:pipeline}).
As mentioned before, we freeze the pre-trained weights of Diff-AE to preserve global semantics and enable Exp-FaceNet to learn explicit face control in a disentangled way. 
However, since Diff-AE backbone already allows a deterministic image reconstruction under this setting, only limited gradients can be generated during error back-propagation, which are far from sufficient to effectively train Exp-FaceNet.
Consequently, it is infeasible to train Exp-FaceNet in a traditional conditional DDIM generation form.
To address this issue, we design a random semantic masking (RSM) strategy, not for the purpose of representation learning like Masked Autoencoders (MAE) \cite{he2022masked}, but rather to achieve the training of Exp-FaceNet effectively.
Concretely, we divide the input image $x_{0}$ into regular non-overlapping patches and randomly mask different portions of patches to obtain a masked image $x_{0}^{m}$ at different timesteps.
By means of this, the semantic latent code $z^m$ encoded by the semantic encoder $\mathcal{E}_\eta$ of Diff-AE only contains fragmented and incomplete content and spatial information of $x_0$.
Meanwhile, the spatial-condition features $f$ generated by Exp-FaceNet $\mathcal{F}_{\phi}$ comprise the fine-grained face shape as well as the camera parameter and lighting condition of $x_0$, which can help to restore the masked face regions in $x_0^m$ in each random denoising timestep. 
The overall training objective can thereby be parameterized as:
\begin{equation}
\mathcal{L}=\mathbb{E}_{x_0, x_{0}^{m},t,\epsilon}[\|\epsilon-\epsilon_\theta(x_t, t, z^m, f)\|_2^2]
\end{equation}
where $\epsilon_\theta$ is the U-Net of Diff-AE which predicts the noise $\epsilon\sim\mathcal{N}(0, \mathbf{I})$ added in noisy image $x_t$.
Throughout the generalized training,  we only train $\mathcal{F}_{\phi}$ and freeze $\epsilon_\theta$ and $\mathcal{E}_{\eta}$ with the pre-trained weights.

\subsection{Exploiting One-shot Semantic Priors}
\label{subsec:one-shot finetuning}
At this point, the well-trained Exp-FaceNet is able to explicitly change pose, expression, and lighting of a facial image.
However, there still exists identity shift or background changes during the face editing, which is especially evident when the target face for editing lies outside the domain of the pre-trained Diff-AE.
Given this, it is meaningful to fully exploit the semantic priors of the to-be-edit image and inject them into the editing process.
Concretely, in this stage, we freeze Exp-FaceNet and only fine-tune Diff-AE with the input portrait image using the aforementioned RSM training strategy.
As a result, this one-shot fine-tuning enables the model to faithfully restore the personalized appearance details as well as editing-irrelevant factors such as background and accessories when performing explicit face editing.

\subsection{Inference Editing}
In practice, we first use EMOCA to predict all explicit parameters (mentioned in Sec. \ref{subsec:Exp-FaceNet}) of the input portrait $x_{0}$, then we modify the pose parameters $\theta$, expression parameter $\psi$, and SH light parameter $l$ by manually setting target values or directly transferring these parameters from a driving portrait.
After this, we calculate the rendered shading shape snapshot $s$ and surface normal snapshot $n$ based on the modified parameters, and further generate explicit control conditions using Exp-FaceNet.
On the other hand, since we should keep a consistent generative mechanisms in training and inference, here we also utilized masked input images to provide high-level semantic conditioning for DDIM decoder and accordingly design an intuitive patch masking strategy for inference editing.
Concretely, for each timestep $t$, we generate the masked image $x^{m}_{t}$ by randomly masking the patches of the input image $x_{0}$ with a linear ratio $\rho_t=0.75-0.5(T-t)/T$, where the number of the inference denoising steps $T$ is set to 20 in this work.
Under this setting, we can generate $x^{m}_{t}$ with high masking ratios $\rho_t$ to emphasize the facial control of the intermediate denoising result $z_{t-1}$ in the early stage of the inference, and then gradually decrease $\rho_t$ to recover semantic information.

%% file: figs/one_shot.tex
\begin{figure*}[t]
    \centering
    \includegraphics[width=1\linewidth]{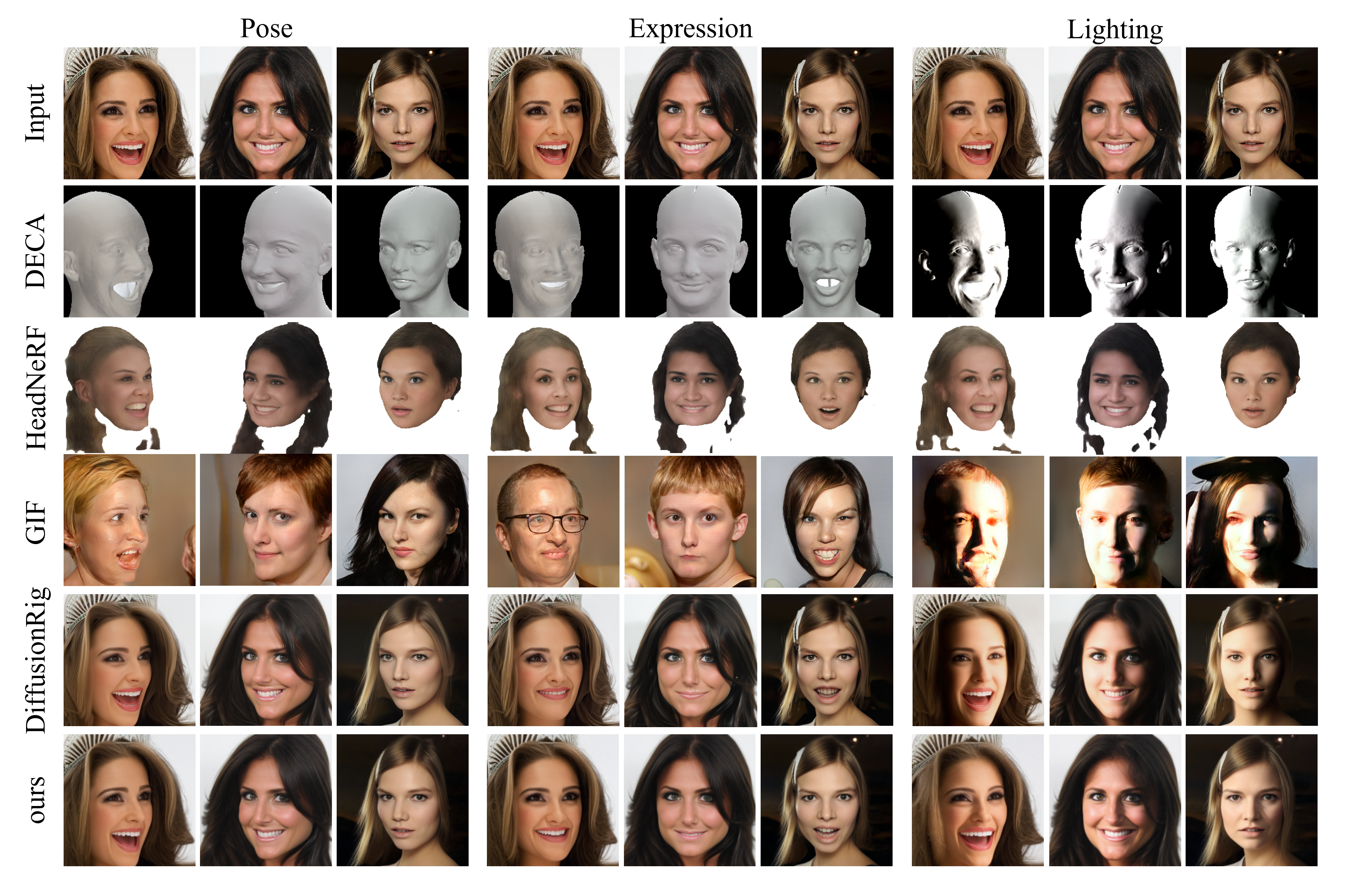}   \caption{\textbf{Qualitative comparison against baselines in one-shot editing.} For each selected image, we use EMOCA \cite{danvevcek2022emoca} to estimate the corresponding explicit parameters, then synthesize the edited images using different methods based on the modified parameters of pose, expression, and lighting. We additionally provide the rendered shading shapes in the second row as the references of explicit control conditions. As can be seen, our DisControlFace can edit images that match well with the target control conditions while faithfully synthesizing facial appearances and editing-irrelevant details.}
    \label{fig:one-shot}
\end{figure*}

%% file: 05_exp.tex
\begin{table}
    \centering
    \begin{tabular}{l@{\hspace{5pt}}c@{\hspace{5pt}}c@{\hspace{5pt}}c@{\hspace{5pt}}c@{\hspace{5pt}}c@{\hspace{5pt}}}
    \hline
    &  ID $\uparrow$  &Shape$\downarrow$&  Pose $\downarrow$ &  Exp $\downarrow$ &Light $\downarrow$ \\
    \hline
 GIF \cite{ghosh2020gif}                                                &  0.22& 3.0& 5.6 & 5.0&0.40\\
 DiffusionRig \cite{ding2023diffusionrig}             & 0.24& 4.3& 4.2& 2.8&0.36\\
 Ours& 0.31&2.8& 4.5& 2.9&0.31\\
     \hline
    \end{tabular}
    \caption{\textbf{Quantitative comparisons against compared baselines} using identity consistency (ID) and DECA re-inference errors on shape, pose, expression, and lighting.}
    \label{tab:quantitative results}
    \vspace{-0.5cm}
\end{table}

\section{Experiments}
\label{sec:Experiments}
\subsection{Implementation Details}
\label{subsec:implem}
We train the proposed Exp-FaceNet on the FFHQ dataset \cite{karras2019style}, which consists of 70k in-the-wild facial images.
For evaluations, we select the images of the CelebA-HQ dataset \cite{karras2018progressive} to perform one-shot fine-tuning and face editing.
To balance generation quality and computational cost, we resize the images to a resolution of $256\times256$ for both training and inference.
Accordingly, we utilize Diff-AE\footnote{https://github.com/phizaz/diffae.} pre-trained on FFHQ-256 for all experiments.
We train Exp-FaceNet for 437,500 iterations, with a learning rate of $1e^{-4}$ and a batch size of 32, while during the one-shot fine-tuning stage, we only fine-tune the pre-trained Diff-AE for 1,500 iterations, with a learning rate of $1e^{-5}$ and a batch size of 4.
In all training stages, we use AdamW \cite{loshchilov2018decoupled} as the optimizer and set the denoising timesteps to 1,000.

\subsection{Comparison}
\label{subsec5.1}
\noindent\textbf{Baselines.}\quad We compare our methods against three generative methods for parametric face image synthesis and editing: HeadNeRF \cite{hong2022headnerf}, GIF \cite{ghosh2020gif}, and DiffusionRig \cite{ding2023diffusionrig}.
Among these, HeadNeRF is a NeRF-based head model, while GIF and DiffusionRig are both generative face models built upon GAN and diffusion model, respectively.
For a fair comparison, we utilize the models pre-trained on FFHQ at a resolution of $256^2$ for each method.

\noindent\textbf{Qualitative comparison.}\quad
We evaluate our DisControlFace against baselines using images from CelebA-HQ and we perform one-shot fine-tuning for all methods.
The qualitative comparison results are provided in Figure \ref{fig:one-shot}.
For all methods, we visualize the editing results of pose, expression, and lighting on three identities.
In order to intuitively measure the editing performance, we further give the rendered snapshot $s$ of the shading shape corresponding to the modified explicit parameters as the references of explicit control conditions.  (second row of Figure \ref{fig:one-shot}).
We can find that compared to the other methods, our method synthesizes images with overall best identity consistency and parametric editing accuracy.
Specifically, HeadNeRF cannot generate realistic facial appearance as well as the background of the original image.
GIF has a good control ability of parametric face editing, however, is completely unable to preserve the face identity.
DiffusionRig, another diffusion-based method achieves better editing results than HeadNeRF and GIF, especially in identity preservation, which can also be attributed to using one-shot fine-tuning to boost the diffusion model with more personal facial priors.
Nevertheless, since DiffusionRig trains the whole model together from scratch which prevents the model from learning robust disentangled control, the edited results still have visible identity shift (\textit{e.g.,} eyes, skin color, and hair) especially in lighting editing.
\\ 

\noindent\textbf{Quantitative comparisons.}\quad
Table \ref{tab:quantitative results} provides the quantitative editing results of all methods on 1000 in-domain images of FFHQ.
Following previous work \cite{ghosh2020gif,deng2020disentangled,ding2023diffusionrig}, we apply DECA re-inference on edited images and calculate the Root Mean Square Error (RMSE) between the input and re-inferred face vertices as well as spherical harmonics to evaluate the editing accuracy on shape, pose, expression, and lighting.
We additionally measure the identity consistency (ID) score by computing the cosine similarity between the deep features generated by ArcFace \cite{deng2019arcface} of the original and edited images.
The results indicate that our method achieve the overall best performance on all metrics.
Note that DisControlFace outperforms other methods by a large margin in ID score, which demonstrates the superiority of our method in identity preservation during one-shot editing.

\input{figs/ablation1}
\input{figs/ablation2}
\subsection{Ablation Study}
\label{subsec5.2}
\noindent\textbf{One-shot fine-tuning.} \quad
Based on the deterministic semantic reconstruction capability of Diff-AE \cite{preechakul2022diffusion}, DisControlFace can extract global semantics of the input face image.
However, since the target portrait image tends to have different face priors with pre-trained Diff-AE, there still exists semantic information shift under the zero-shot editing scenario, as shown in Figure \ref{fig:ab1}.
Given this, we adopt a simple one-shot fine-tuning on the pre-trained Diff-AE to fully exploit the semantic priors of the to-be-edit image and inject them into the editing process.
We can observe that by means of this, the face identity and other high-level semantic information (\textit{e.g.,} background, accessories, and hair) can be well preserved.\\

\input{figs/ablation34}
\noindent\textbf{Effectiveness of RSM training.} \quad
In this paper, we achieve a disentangled training of the proposed Exp-FaceNet by freezing the pre-trained weights of Diff-AE and according designing a RSM training strategy.
To demonstrate the necessity of the proposed RSM training, we separately train Exp-FaceNet with and without random patch masking for the input image of semantic encoder of the Diff-AE backbone.  
Figure \ref{fig:ab2} shows that both training strategies enable the model to reconstruct the input image. 
However, only the model trained with RSM strategy can generate images with novel poses. 
This result is consistent with our claim that since the pre-trained Diff-AE backbone can already allow deterministic image reconstruction, limited gradients can be generated during error back-propagation for an effective training of Exp-FaceNet.\\

\noindent\textbf{Effectiveness of our disentangled pipeline}
To further demonstrate the advantages of our method over traditional pipeline in terms of disentangled control, we train the whole model (Diff-AE+Exp-FaceNet) from scratch without using RSM training.
The result in Fig. \ref{fig:ab34} (a) shows that our disentangled pipeline can significantly reduce identity shift, both visually and in terms of quantitative metrics.\\

\input{figs/disentangled_control}
\noindent\textbf{Exp-FaceNet structure.} \quad
Our Exp-FaceNet is inspired from ControlNet \cite{zhang2023adding}, a popular deep network which has been widely employed to add various spatial visual guidance (\textit{e.g.,} edge maps, pose maps, depth maps, \textit{etc}.) to Stable Diffusion (SD) \cite{rombach2022high} for text-to-image generation.
However, there exists many differences between two models. 
First, ControlNet is specially designed for SD which can generate specific content based on the input prompts, while there still exists uncertainty and randomness in the generation controlled by visual guidance and text prompts.
In contrast to this, this paper focuses on semantics preservation and explicit editing of the input portrait image.
Given this, our Exp-FaceNet is designed to be compatible with a Diff-AE backbone, aiming to provide semantically deterministic DDIM conditioning.
Second, we construct Exp-FaceNet as a U-Net instead of adopting zero convolutions introduced in ControlNet.
With this setting, we can endow Exp-FaceNet with strong capability of U-shape models in extracting spatial deep features from pixel-aligned visual guidance map (rendered explicit snapshots in this work).
Here we compare the generation performance between using Exp-FaceNet and ControlNet to learn explicit face control.
The results in Fig. \ref{fig:ab34} (b) show that compared to ControlNet, our Exp-FaceNet can help to edit the face image with less DECA re-inference error and visualized better identity consistency.

\subsection{Visualization of Disentangled Control.}
To further analyze the disentangled conditioning of our DisControlFace, we show the synthesis results of mixed conditioning generation in Figure \ref{fig:discontrol}.
Specifically, we use the images in the blue column to perform one-shot fine-tuning for Diff-AE, which helps to restore the face identity and other high-level semantic information of these images.
For explicit conditioning, considering that the 3DMM parameters of shape, texture, and person-specific details are related to face ID, we extract all the EMOCA-estimated 3DMM parameters of the images in the blue column but only replacing the parameters of pose, expression, and lighting with those estimated from the images in the right row.
We can observe that each synthesis face has the same face shape, pose, expression, and lighting condition with the corresponding image in the right row.
Meanwhile, all facial appearance priors like skin color, eyes, and lips color as well as non-facial high-level semantics such as background and hair of the corresponding image in the blue column have been successfully preserved in the synthesized image.
This result can intuitively show the effectiveness of the disentangled face control capability of DisControlFace.
Also benefiting from this, the proposed DisControlFace can preserve original high-level semantics while performing fine-grained explicit face editing.

\subsection{Applications}
\noindent\textbf{Semantic manipulation.}\quad
Since we adopt Diff-AE \cite{preechakul2022diffusion} as the reconstruction backbone and freeze the pre-trained weights without tuning, our DisControlFace inherits the encoding capability of global facial semantics.
Therefore, compared to previous diffusion-based model such as DiffusionRig \cite{ding2023diffusionrig}, our model also supports manipulating face attributes (\textit{e.g.,} hairstyle, age, and accessories) of the input portrait by linearly editing global semantic codes.
The visualized results of both semantic manipulation and explicit editing are shown in Figure \ref{fig:implicit_explicit} , which once again demonstrate the effectiveness and flexibility of the proposed disentangled conditional generation mechanism.\\

\input{figs/implicit_explicit}
\input{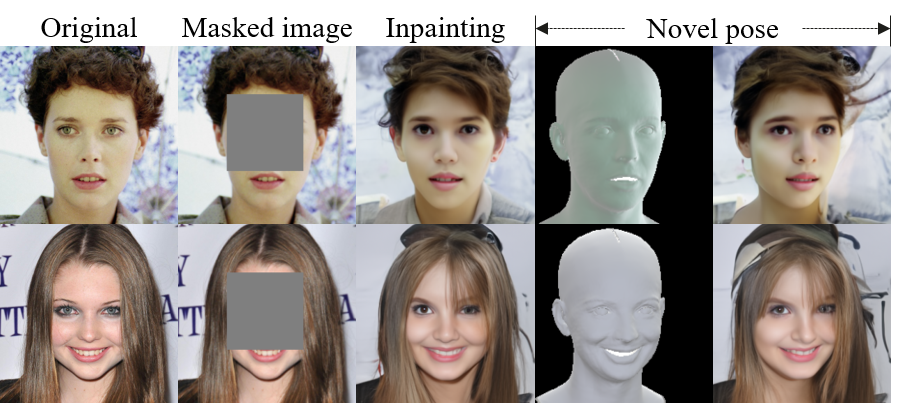}
\noindent\textbf{Image inpainting.}\quad
Benefiting from the proposed RSM training strategy, our model inherently supports image inpainting. 
Figure \ref{fig:inpainting} shows the results of zero-shot inpainting on center-masked facial images as well as the subsequent editing.
We can observed that the restored face in the inpainted image is smooth and natural, also has good similarity with the original face.
Besides, the explicitly edited image still shows a consistent identity with the restored facial image.
Meanwhile, since it is inappropriate to use masked images to perform one-shot fine-tuning on Diff-AE, some editing-irrelevant high-level semantics such as hairstyle and background might not be well preserved in 
generation. 

%% file: figs/ablation1.tex
\begin{figure}[t]
    \centering
    \includegraphics[width=1\linewidth]{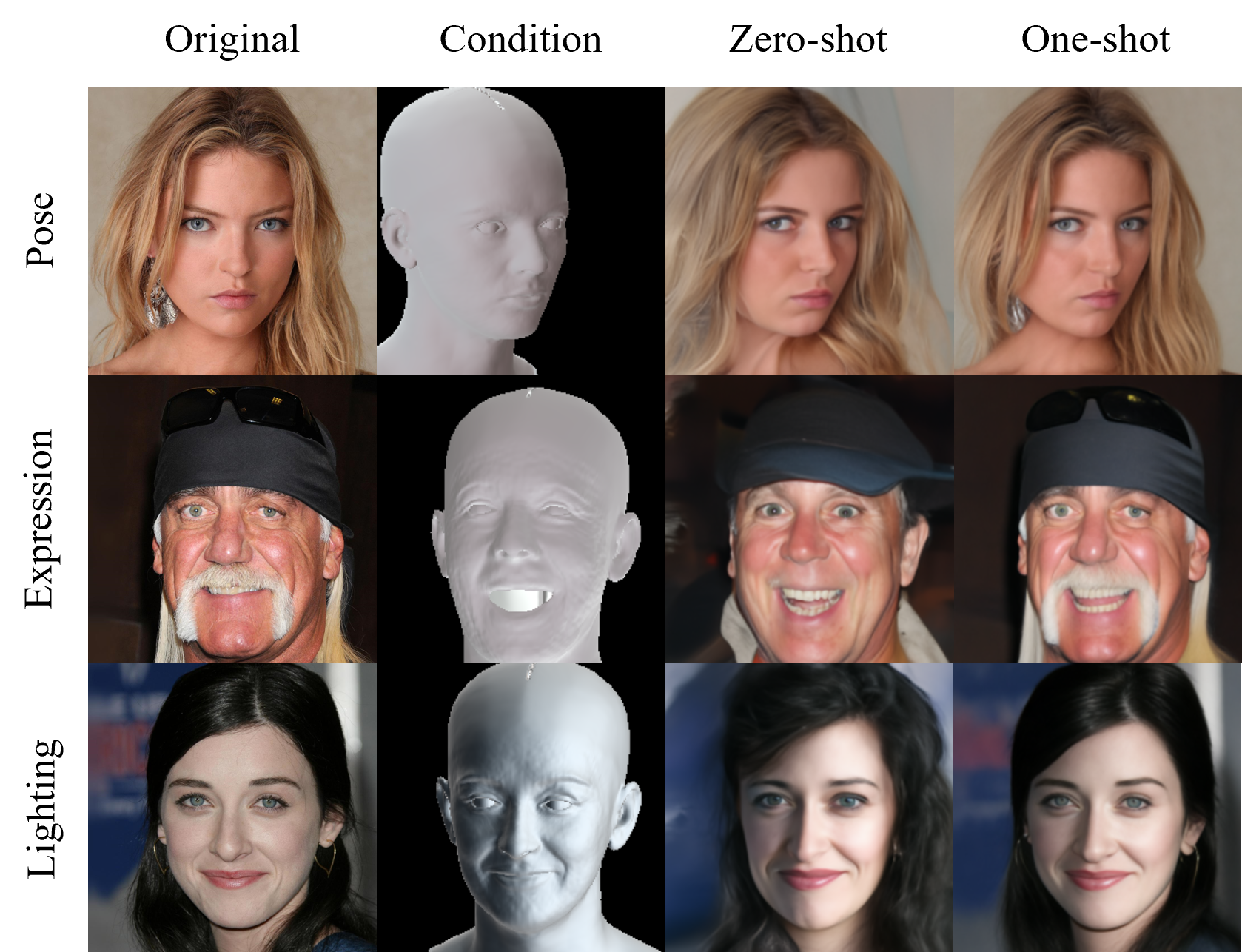}   \caption{Ablation study on one-shot fine-tuning. DisControlFace can perform zero-shot explicit editing, where however, identity shift still exists. On this basis, adopting a simple one-shot fine-tuning can significantly improve the preservation of face identity as well as other editing-irrelevant semantic information.}
    \label{fig:ab1}
\end{figure}

%% file: figs/ablation2.tex
\begin{figure}[t]
    \centering
    \includegraphics[width=1\linewidth]{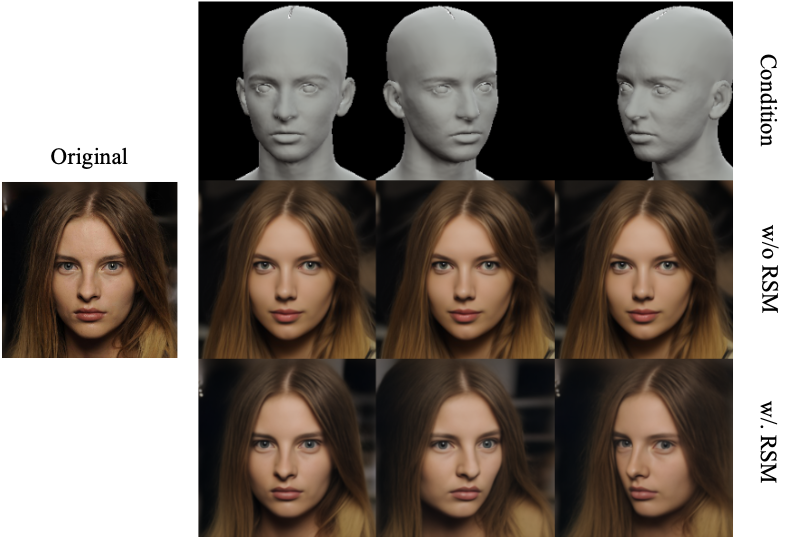}   \caption{The necessity of the proposed random semantic masking (RSM) training. Without RSM training, it is infeasible to train Exp-FaceNet with explicit face control. }
    \label{fig:ab2}
\end{figure}

%% file: figs/ablation34.tex
\begin{figure}[t]
    \centering
    \includegraphics[width=1\linewidth]{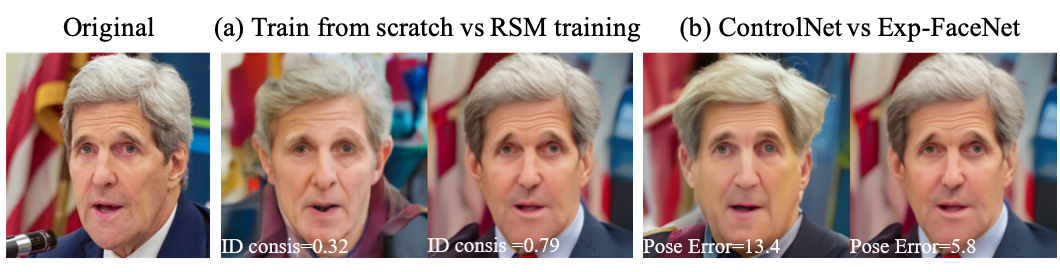}   \caption{The ablation studies on our disentangled pipeline (a) and Exp-FaceNet structure (b). The disentangled control setting in DisControlFace trained with RSM strategy can significantly improve the identity preservation in explicit editing. Compared to adopting ControlNet with a light-weight decoder and zero convolutions, using our Exp-FaceNet can improve the explicit control accuracy by a large margin.}
    \label{fig:ab34}
\end{figure}

%% file: figs/disentangled_control.tex
\begin{figure}[t]
    \centering
    \includegraphics[width=1\linewidth]{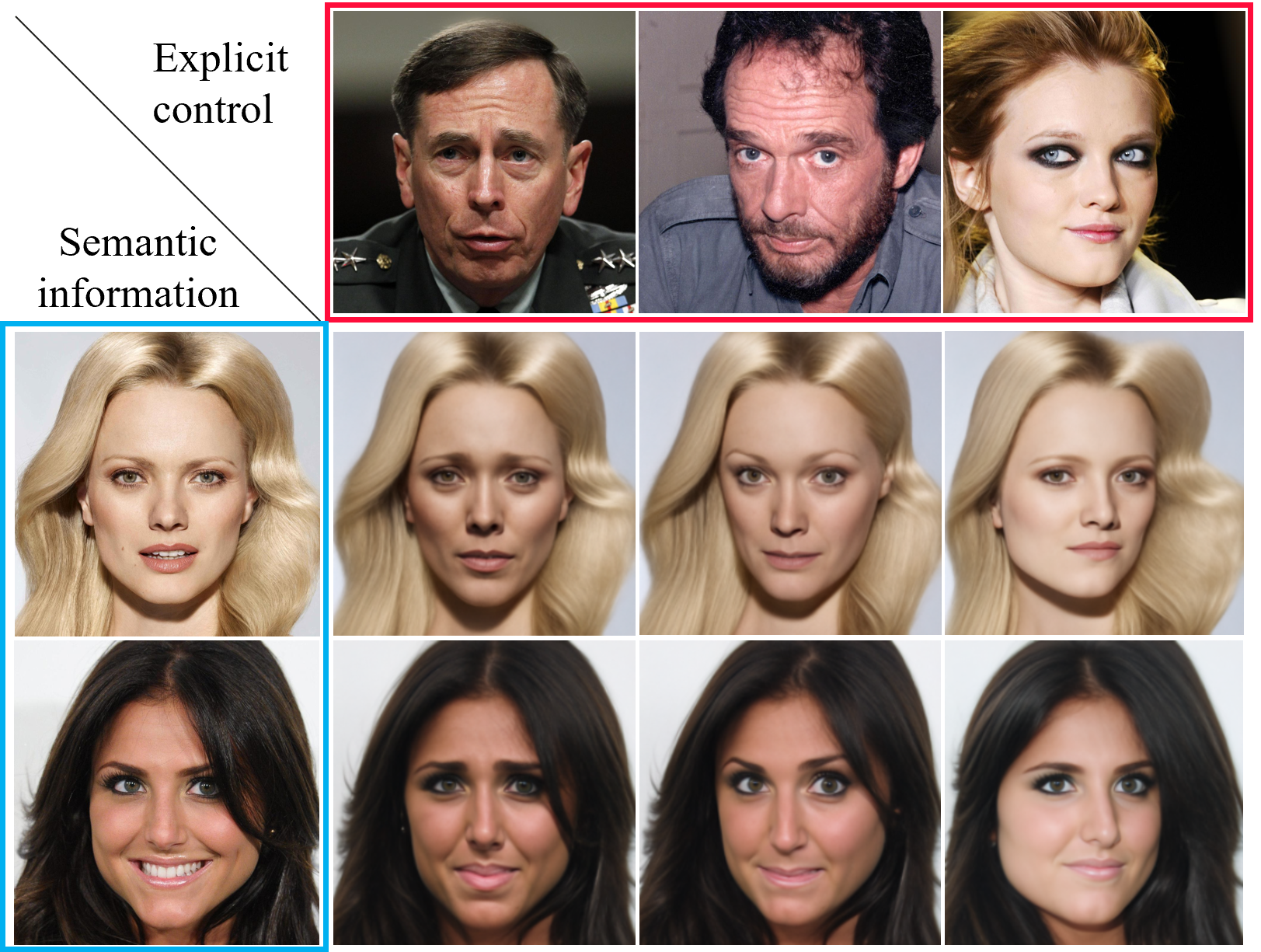}   \caption{Visualization of disentangled face control. We separately utilize the encoded global semantic code of one image and the estimated 3DMM parameters of another image to provide semantic conditioning and explicit parametric conditioning in facial image generation.}
    \label{fig:discontrol}
    \vspace{-0.5cm}
\end{figure}

%% file: figs/implicit_explicit.tex
\begin{figure}[t]
    \centering
    \includegraphics[width=1\linewidth]{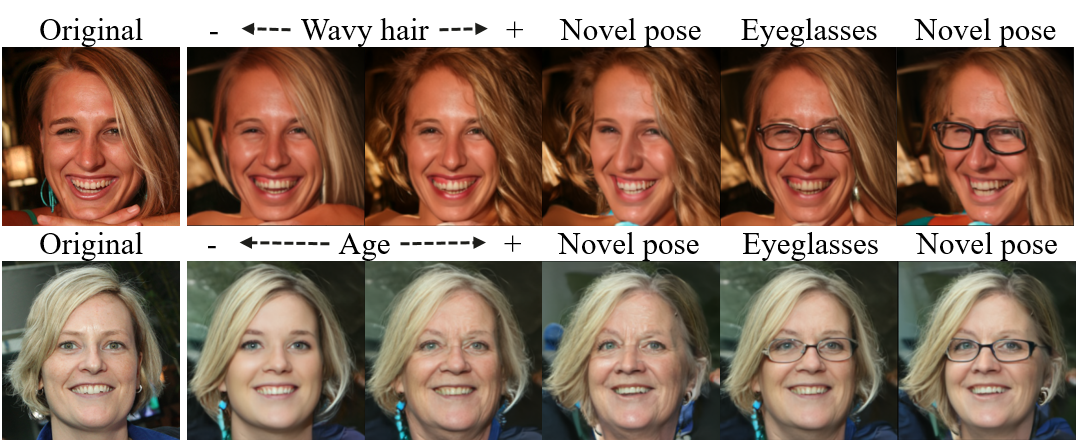}   \caption{\textbf{Synthesis results in semantic manipulation.} We follow Diff-AE \cite{preechakul2022diffusion} to manipulate the global facial attributes (age, hairstyle, and eyeglasses) by linearly editing $z$. Owing to the disentangled control mechanism, our model can simultaneously perform semantic manipulation and explicit editing of the input face image.}
    \label{fig:implicit_explicit}
\end{figure}

%% file: figs/inpainting.tex
\begin{figure}[t]
    \centering
    \includegraphics[width=\linewidth]{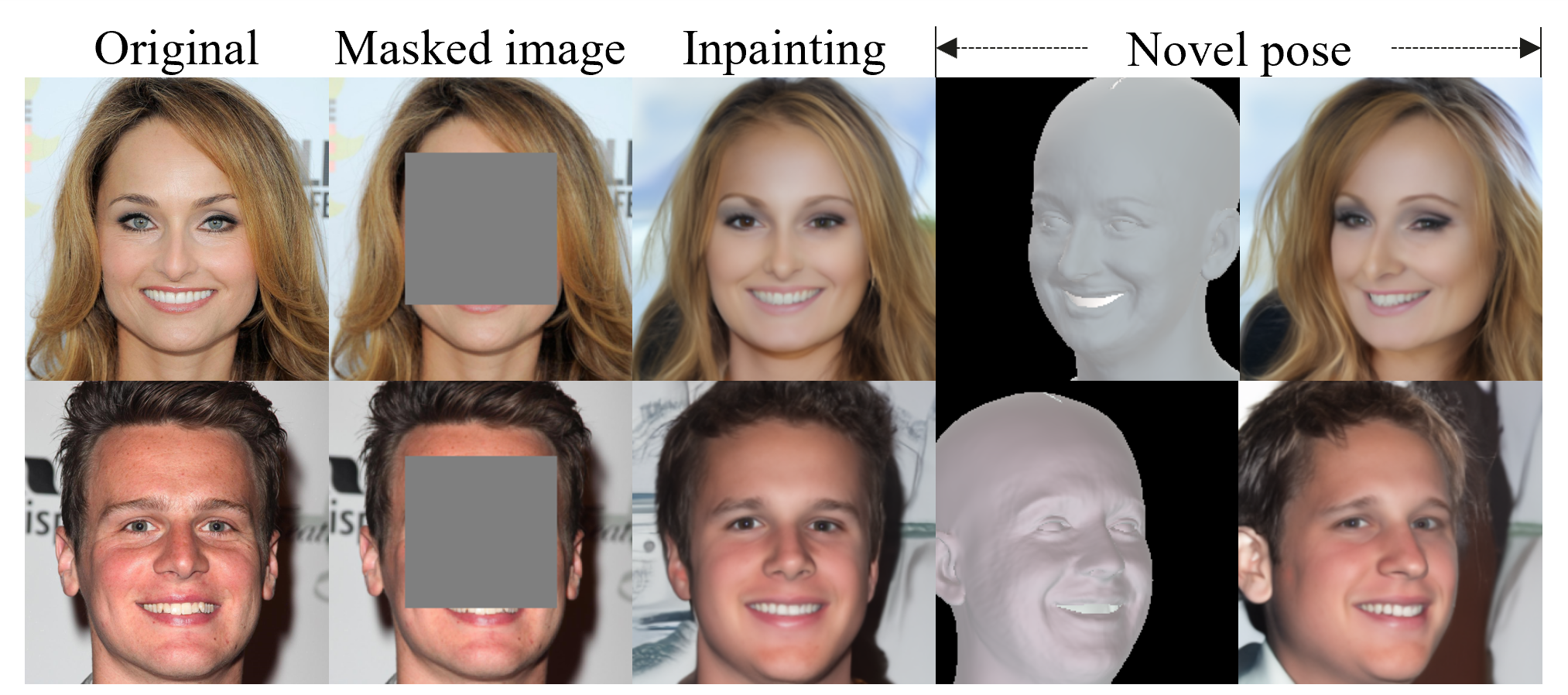}   
    \caption{\textbf{Zero-shot inpainting and subsequent explicit editing} on images from CelebA-HQ. DisControlFace can restore the masked face regions smoothly and naturally. On this basis, Our method can further edit the restored facial images based on the modified explicit 3DMM parameters.}
    \label{fig:inpainting}
\end{figure}

%% file: 06_conclusion.tex
\section{Limitations and Conclusion}
\label{sec:conclusion}
\noindent\textbf{Limitations and future work.}\quad 
In this work, both adopted Diff-AE and constructed Exp-FaceNet were trained using the FFHQ dataset, which consisting 70,000 in-the-wild images.
However, this data size is still not sufficient to train the model to learn a more generalized face priors.
It can be expected that collecting much more face data with abundant face conditions (\textit{e.g.,} pose and expression) for training can substantially improve the editing performance, especially for some challenging tasks like zero-shot explicit editing.
We adopt EMOCA to estimate explicit face parameters of the input image and generate spatial-aware conditions representing control information based on them.
However, EMOCA struggles to model the detailed geometry of eyeballs as well as some extreme expressions, which could hinder the model from restoring the corresponding facial details when generating edited images.
Furthermore, our editing framework is constructed as a denoising diffusion pipeline, which therefore is still unable to compete with GANs in terms of generation speed.
In the future, with the development of fast sampling algorithms, we expect the generation time of our model to further decrease.\\

\noindent\textbf{Conclusion.}\quad
We have presented DisControlFace, a novel diffusion framework for one-shot facial image editing.
Through exploiting disentangled conditioning on high-level semantics and explicit 3DMM parameters in the generation process, our model excels in explicit and fine-grained face control while preserving semantic information and facial priors in face editing.
This may boost a series of related applications including various semantic and explicit facial image editing, zero-shot image inpainting, and cross-identity face driving.

%% file: 07_ackn.tex
\section{Acknowledgements}
This work was supported in part by the National Natural Science Foundation of China under Grant 62271405, the National Natural Science Foundation of China under Grant 62171255, and the Tsinghua University - Migu Xinkong Culture Technology (Xiamen) Co.Ltd. Joint Research Center for Intelligent Light Field and Interaction Technology.

%% file: 08_appendix.tex
\setcounter{section}{0}
\setcounter{figure}{0}    
\setcounter{table}{0}

\section{Additional Implementation Details}
\label{implem}
The complete configurations of the training of Exp-FaceNet and one-shot fine-tuning are shown in Table \ref{tab:training_configuration}.
We utilized the official codes to generate the results of the compared baselines HeadNeRF\footnote{https://github.com/CrisHY1995/headnerf} \cite{hong2022headnerf}, GIF\footnote{https://github.com/ParthaEth/GIF} \cite{bergman2022generative}, and DiffusionRig\footnote{https://github.com/adobe-research/diffusion-rig} \cite{ding2023diffusionrig}.
In inference, we generate the initial noise map $X_T$ by sampling from $\mathcal{N}(0, I)$ rather than computing through a reverse deterministic generative process proposed in the original Diff-AE \cite{preechakul2022diffusion}.
The further analysis of this setting is provided in Section \ref{Semantic Conditioning of Diff-AE}.

\begin{table}
    \centering
    \begin{tabular}{lcc}
    \hline
         &  Exp-FaceNet training& Fine-tuning\\
         \hline
 Image size& \multicolumn{2}{c}{256}\\
 Patch size&  \multicolumn{2}{c}{16}\\
 Normalization& \multicolumn{2}{c}{[-1, 1]}\\
 Masking ratio& \multicolumn{2}{c}{sample from $U(0.25, 0.75)$}\\
 Optimizer& \multicolumn{2}{c}{AdamW (no weight decay)}\\
 Diffusion loss     & \multicolumn{2}{c}{MSE loss}\\
 EMA decay& 0.9999&$\setminus$\\
 Learning rate&  1e-4& 1e-5\\
 Batch size&  32& 4\\
 Denoising steps & \multicolumn{2}{c}{1000}\\
 Iterations &  437500 & 1500\\
 Device& 8 V100s&1 V100\\
 Training time &3 days & 4minutes\\
    \hline
    \end{tabular}
    \caption{\textbf{The complete configurations of the training of Exp-FaceNet and one-shot fine-tuning.}}
    \label{tab:training_configuration}
\end{table}

\section{Additional Experiments and Analyses}
\label{add_exp}

\input{figs/diff-ae}
\subsection{Semantic Conditioning of Diff-AE}
\label{Semantic Conditioning of Diff-AE}
In original Diff-AE \cite{preechakul2022diffusion}, the initial noise map $X_T$ in inference is computed through a reverse deterministic generative process, which has a capacity for capturing stochastic details.
In Figure \ref{fig:diff_ae}, we compare the reconstruction and editing results of separately using the reverse deterministic noise and the randomly sampled noise as the initial denoising map in Diff-AE. 
It can be observed that using a deterministic initial noise map can achieve a better reconstruction of the input image such as the background, however generating the editing images with less accurate explicit control and incoherent facial appearance.
This might because that the reverse deterministic noise computing can encode the stochastic details of the input image, which is crucial for a near-exact reconstruction.
However, this process inherently conflicts with the explicit face editing where some details of the input image should be changed based on the modifications of the explicit parameters.
Also it is inconsistent with the initialization in the training of Exp-FaceNet.
Given this, in this work, we choose to generate the initial noise map $X_T$ by sampling from $\mathcal{N}(0, I)$ in inference.
Moreover, we can find that using a simple one-shot fine-tuning can effectively enhance the semantics preservation in the editing result, \textit{e.g.,} more faithful hair style.

\input{figs/sup_ab2}
\input{figs/inference}
\subsection{Additional Ablation Studies}
\noindent\textbf{Fine-grained face geometry.}\quad
In EMOCA \cite{danvevcek2022emoca}, a person-specific detail vector $\delta$ is specially estimated and can be further combined with pose parameter $\theta$ and expression parameter $\phi$ to generate the expression dependent displacement map for refining the face geometry with animatable wrinkle details.
Here we compare the editing results between using estimated FLAME parameters to calculate the 3D face mesh and additionally introducing the detail vector $\delta$ in the mesh calculation.
Figure \ref{fig:sup_ab2} demonstrates that using $\delta$ can generate a rendered snapshot with refined expression-dependent face geometry, which thereby helps to recover the detailed expressions (\textit{e.g.,} wrinkles) of the edited face.\\

\noindent\textbf{Different masking strategies in inference.}\quad
Here we explore how the masking strategy used in inference affects the editing performance.
The comparison is shown in Figure \ref{fig:sup_ab1}, we can observe that when setting masking ratio to 0\% for all inference steps, the edited image can not match the control signal very well which can be attributed to strong deterministic reconstruction in this setting.
Meanwhile, setting masking ratio to 75\% and 25\% for all inference steps slightly harm the semantics recovering and explicit control, respectively.
Besides, we can see the other three masking strategies can achieve better editing results where the proposed linear masking ratio can perform overall best editing with accurate face control and good preservation of facial semantics.\\

\section{Model Architectures}
\label{model}
The structural details of the proposed DisControlFace is presented in Figure \ref{fig:structure}.
The detailed architecture of Diff-AE is provided in the published paper \cite{preechakul2022diffusion} and released code\footnote{https://github.com/phizaz/diffae}.
The proposed Exp-FaceNet mirrors the structure of the Conditional DDIM (\textit{i.e.,} U-Net) in Diff-AE, which however, customizes the input layer by setting the input channel number to 6, allowing it to take the concatenated snapshots as input.
On this basis, we fuse the 2D feature maps outputted by the input layers of Conditional DDIM and Exp-FaceNet by pixel-wise summation, then feeding the fused feature maps into the subsequent layers of Exp-FaceNet for generating spatial-wise explicit conditioning features.
To further provide fine-grained conditioning for the diffusion generation process, we add multi-scale features outputted by the decoder blocks of the Exp-FaceNet ($f_A$ to $f_M$ in Figure \ref{fig:structure}) back to the corresponding blocks of the Diff-AE backbone.
\input{figs/structure}

\section{Additional Limitations and Future Work}
\label{discuss}
Our DisControlFace has a separate editing control network besides the U-Net noise predictor and performs denoising diffusion process in image space, which results in the model being able to generate images with limited resolutions.
Potential future improvements includes introducing a light-weight super-resolution network to the model or extending the model to a latent diffusion version.

In the proposed Masked Diff-AE training, we randomly mask some patches of the input image as the input of the semantic encoder of the Diff-AE backbone, which enables an effective training of Exp-FaceNet in a disentangled setting.
Meanwhile, only performing random masking on face-related patches is expected to further improve the consistency of the background region of the input portrait with different editing applied, which however might slightly increase the training and inference time.
Corresponding explorations can serve as another potential valuable future work.

Moreover, currently our framework is still hard to perform reliable one-shot explicit face editing on profile face images. 
This limitation arises for two main reasons. 
First, the existing public 2D in-the-wild portrait datasets, such as FFHQ and CelebA-HQ, contain a limited number of profile face images. 
Consequently, it is challenging to train the face model with sufficient prior knowledge of profile faces using these datasets. Second, our framework employs a 3D face model to estimate explicit 3DMM parameters for subsequent explicit face editing. 
However, existing 3D face models, \textit{e.g.,} DECA and EMOCA, also encounter difficulties in accurately estimating the parameters of profile faces.
Despite these, we believe that by collecting more profile face images to enrich the training data and developing more robust 3D face models, our framework can better support explicit face editing on profile images.


%% file: figs/diff-ae.tex
\begin{figure}[t]
    \centering
    \includegraphics[width=1\linewidth]{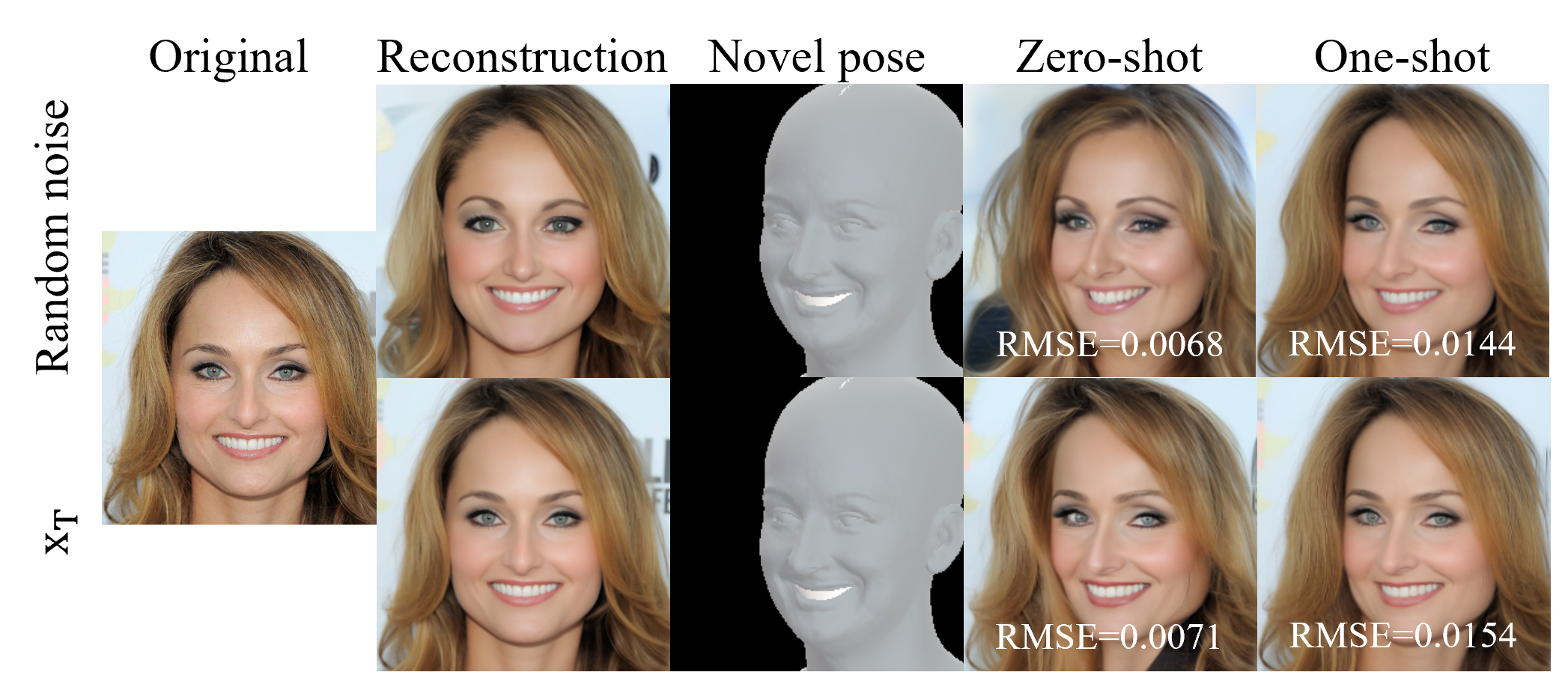}   \caption{The reconstruction and editing results of using different initial noise map computing strategy in Diff-AE. $X_T$ indicates using the original reverse deterministic computing to generate the initial noise map of Diff-AE.}
    \label{fig:diff_ae}
\end{figure}

%% file: figs/sup_ab2.tex
\begin{figure}[t]
    \centering
    \includegraphics[width=1\linewidth]{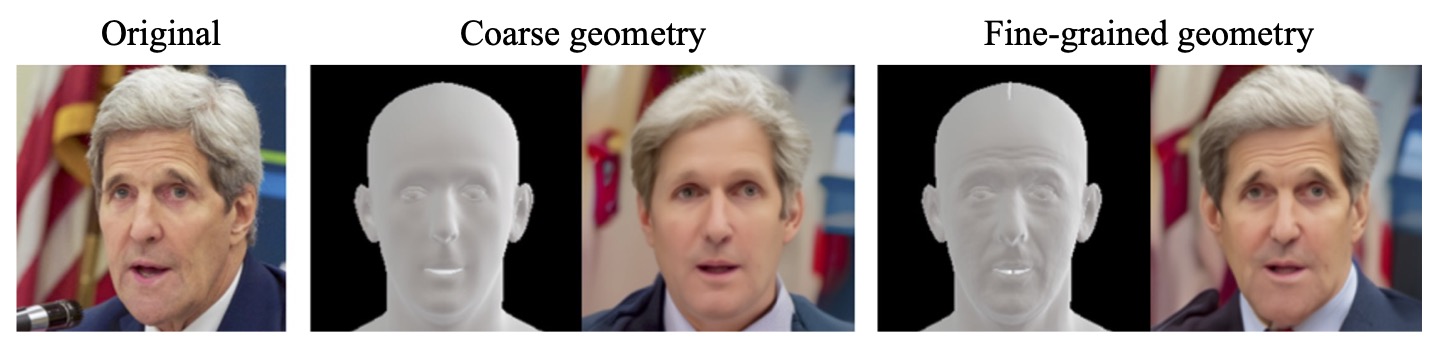}   \caption{Ablation study on fine-grained face parameters. Adopting the detail vector $\theta$ estimated by EMOCA \cite{danvevcek2022emoca} can help to generate the control condition with more fine-grained face geometry, which allows faithful facial details preservation.}
    \label{fig:sup_ab2}
\end{figure}

%% file: figs/inference.tex
\begin{figure*}[t]
    \centering
    \includegraphics[width=1\linewidth]{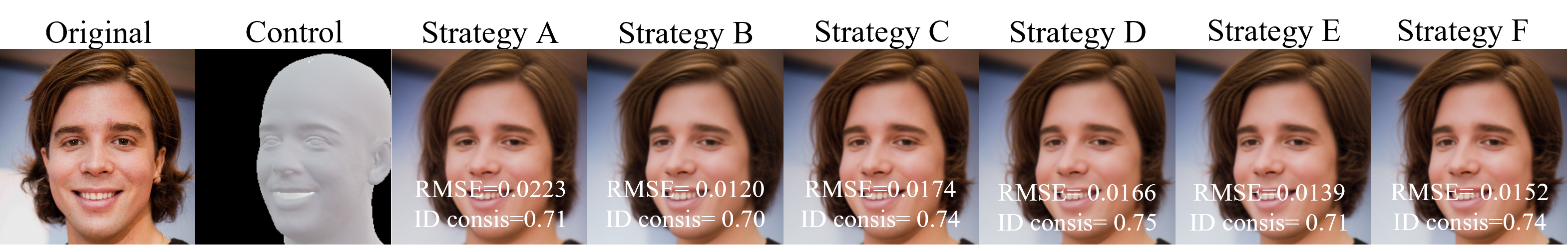}   \caption{Ablation study on different masking strategies in inference. We set the inference denoising steps to 20 for all masking strategies. Strategy A: the masing ratio is set to 0\% for all 20 steps; Strategy B: the masking ratio is set to 75\% for all 20 steps; Strategy C: the masking ratio is set to 25\% for all 20 steps; Strategy D: the masking ratio is set to 25\% and 75\% for the first 10 steps and last 10 steps; Strategy E: the masking ratio is set to 75\% and 25\% for the first 10 steps and last 10 steps: Strategy F: the linear masking ratio introduced in the main paper.}
    \label{fig:sup_ab1}
\end{figure*}

%% file: figs/structure.tex
\begin{figure*}[t]
    \centering
    \includegraphics[width=0.85\linewidth]{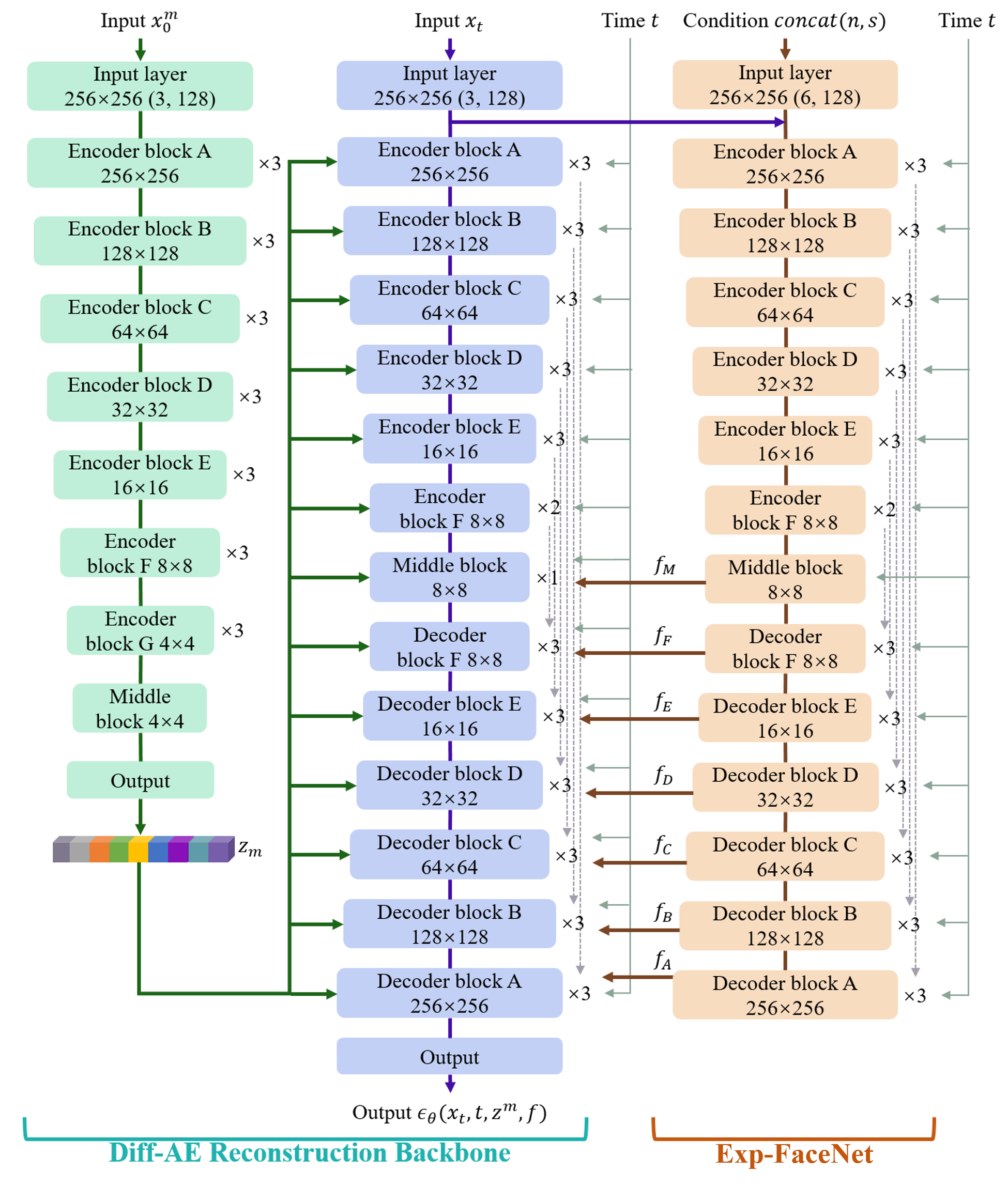}   \caption{\textbf{Detailed architecture of the proposed DisControlFace.} Please zoom in to see details. } 
    \label{fig:structure}
\end{figure*}